\documentclass[runningheads]{llncs}

\usepackage{eccv}

\usepackage{eccvabbrv}
\usepackage{wrapfig}
\usepackage{graphicx}
\usepackage{booktabs}
\usepackage{multirow}
\usepackage{makecell}
\usepackage{adjustbox}
\usepackage[accsupp]{axessibility}  % Improves PDF readability for those with disabilities.

\usepackage{hyperref}

\usepackage{orcidlink}

\begin{document}

\title{BiSLW: Bi-Spectral Latent Watermarking for Generative Diffusion Models} 

\titlerunning{BiSLW: Bi-Spectral Latent Watermarking}

\author{Aryan Pandit\inst{1}\orcidlink{0009-0004-8146-548X}}

\authorrunning{A. Pandit}

\institute{PDPM Indian Institute of Information Technology, Jabalpur, India\\\email{23bec023@iiitdmj.ac.in}}

\maketitle

\begin{abstract}
  Diffusion-based generative models have transformed visual content synthesis, yet they remain vulnerable to unauthorized usage and lack reliable attribution methods. Existing watermarking techniques often treat latent tensors as static spatial feature maps or depend on pixel-domain modification, and most do not explicitly leverage the internal frequency structure of the latent space for dual-band redundant embedding, leaving them susceptible to the stochastic nature of diffusion and regeneration attacks. We introduce BiSLW, a trainable bi-spectral latent watermarking framework that jointly embeds aligned identity signals across complementary spectral bands of the decoded diffusion latent using learned encoders and decoders, going beyond fixed-pattern frequency approaches. We leverage the inherent frequency structure of diffusion latents to design a dual-band watermarking framework. Low-frequency components encode global semantics, while high-frequency components capture fine texture. We exploit this structure to embed watermarks across complementary spectral bands. The watermark is independently injected into both bands via learned encoders and recombined before decoding, ensuring it becomes intrinsic to the generative trajectory. Dual spectral decoders recover the watermark from each band, while a cross-band consistency constraint enforces alignment between semantic and textural embeddings. Experiments show that BiSLW achieves a strong balance between perceptual fidelity and robustness, improving PSNR by over 3 dB compared to prior latent diffusion watermarking methods while preserving near-perfect bit accuracy under aggressive regeneration and common distortions, all with negligible computational overhead. 
\end{abstract}

\section{Introduction}
\label{sec:intro}

Over the past few years, diffusion-based generative models have established themselves as the leading approach for producing photorealistic synthetic images. Advances such as Denoising Diffusion Probabilistic Models (DDPM)~\cite{ho2020denoisingdiffusionprobabilisticmodels}, Denoising Diffusion Implicit Models (DDIM)~\cite{song2022denoisingdiffusionimplicitmodels}, and Latent Diffusion Models (LDM)~\cite{rombach2022highresolutionimagesynthesislatent} enable the generation of highly realistic images across diverse domains. However, as these tools grow more widely available and their generated content becomes increasingly indistinguishable from real-world media, a pressing question arises: how do we trace the origin of such content and hold its creation accountable?

Invisible watermarking has emerged as a compelling solution for embedding verifiable identity markers directly into generated images. Yet, current watermarking techniques designed for diffusion models fall short in several important ways. Methods that operate after image generation --- working directly on pixel-level outputs~\cite{bui2023rostealsrobuststeganographyusing, fernandez2023stablesignaturerootingwatermarks} --- not only add computational burden but are also susceptible to attacks that exploit the diffusion process itself to strip away embedded signals. Latent-based methods improve efficiency by modifying intermediate representations, however, most treat these latent tensors purely as spatial feature maps without explicitly exploiting the frequency structure that organizes them. While some prior work such as Tree-Ring~\cite{wen2023treeringwatermarksfingerprintsdiffusion} does operate in the frequency domain of the initial diffusion noise, it embeds fixed Fourier-domain patterns and relies on inversion-based recovery rather than learning dual-band representations end-to-end. Watermarks anchored to a single representational mode — whether spatial or fixed-frequency --- can degrade or vanish entirely when subjected to regeneration pipelines or distortions in the signal space.

To address these limitations, we present \textbf{BiSLW}, a bi-spectral latent watermarking framework that leverages the inherent frequency hierarchy of diffusion latents: low-frequency components capture global semantic structure while high-frequency components encode fine-grained texture. BiSLW decomposes the latent tensor into these spectral bands, independently injects a shared watermark into each via dedicated encoders, and reassembles them before VAE decoding — ensuring the watermark is intrinsic to the generative representation. A cross-band consistency constraint prevents spectral drift, while dual spectral decoders provide complementary redundancy that strengthens overall robustness.

This bi-spectral embedding strategy extends conventional spatial latent watermarking methods by introducing structured dual-band redundancy. By binding the watermark simultaneously to semantic layout and texture statistics in latent space, BiSLW improves resilience to diffusion regeneration, geometric transformations, compression artifacts, and common signal distortions. Through extensive experimentation, we demonstrate that our approach delivers over 3 dB improvement in perceptual quality relative to strong latent watermarking baselines, all while sustaining near-perfect bit accuracy even under aggressive attack conditions and doing so with negligible overhead introduced by the embedding process.

In summary, our contributions are threefold:
\begin{itemize}
\item We propose a trainable bi-spectral latent watermarking framework that jointly embeds aligned identity signals across complementary semantic (low-frequency) and textural (high-frequency) spectral bands of the decoded diffusion latent, going beyond fixed-pattern frequency approaches by learning dual-band encoders and decoders end-to-end.

\item We present cross-band identity binding for consistent watermark encoding across semantic and textural bands.

\item We demonstrate state-of-the-art robustness and perceptual quality for diffusion watermarking with minimal computational overhead.
\end{itemize}

\section{Related Work}
\label{sec:related}

\textbf{Detection of AI-Generated Images.}
The widespread adoption of diffusion-based generative models has motivated
extensive research on detecting synthetic imagery.
Early efforts in this space focused on training discriminative classifiers to separate generated images from authentic ones, typically by exploiting statistical artifacts or frequency-domain inconsistencies~\cite{corvi2022detectionsyntheticimagesgenerated, wang2020cnngeneratedimagessurprisinglyeasy}. 
Although effective under controlled settings, such detectors often degrade under distribution shifts, post-processing, or emerging architectures. They also provide limited attribution, identifying synthetic content without revealing its source.

\textbf{Image Watermarking.}
Watermarking research spans a broad spectrum of techniques, from hand-crafted spatial and frequency-domain methods to modern learning-based 
frameworks~\cite{cox2007digital}. Frequency-domain approaches have proven particularly effective, leveraging transforms such as DCT and DWT that naturally lend themselves to robustness against compression artifacts and filtering operations~\cite{Yldz2023DigitalIW}.
More recently, deep learning has reshaped the field, with end-to-end frameworks like RivaGAN~\cite{zhang2019robustinvisiblevideowatermarking}, SSL~\cite{fernandez2022watermarkingimagesselfsupervisedlatent}, and FNNS~\cite{sym17071094} learning joint embedding and extraction pipelines that demonstrate strong resilience to a wide variety of common distortions.
However, these methods typically operate as post-processing steps
in the pixel domain and are not designed to account for the stochastic
regeneration process of diffusion models.

\textbf{Watermarking for Generative Models.}
Recent approaches embed watermarking mechanisms directly into generative models. Stable Signature~\cite{fernandez2023stablesignaturerootingwatermarks} embeds signatures within the latent decoder of diffusion models, while other approaches modify intermediate latent representations during generation. Although such methods reduce post-processing overhead, they commonly treat latent tensors purely as spatial feature maps and do not explicitly exploit their internal structure. Consequently, watermark signals may degrade when latent representations are perturbed during sampling or regeneration.

\textbf{Generation-Time Watermarking.}
Another line of work embeds watermarks by modifying the generation process itself. Tree-Ring watermarking~\cite{wen2023treeringwatermarksfingerprintsdiffusion} injects structured circular patterns into the frequency domain of the initial diffusion noise, showing that frequency-domain properties can be leveraged for watermarking — though via fixed patterns and inversion-based recovery rather than learned networks. RoSteALS~\cite{bui2023rostealsrobuststeganographyusing} introduces sampling-time latent embedding. Both approaches tightly couple watermark signals with the diffusion trajectory but rely on specific sampling assumptions or inversion procedures.  Recent work on adapter-based latent embedding~\cite{ci2024wmadapteraddingwatermarkcontrol} and Fourier-domain robustness~\cite{lee2025semanticwatermarkingreinventedenhancing} further explores frequency-aware strategies; we discuss their relationship to BiSLW below.

\textbf{Positioning of BiSLW.}
While Tree-Ring~\cite{wen2023treeringwatermarksfingerprintsdiffusion} already exploits frequency-domain properties of diffusion noise, our contribution is distinct: BiSLW is a trainable dual-band framework performing learned embedding and extraction directly in the decoded latent (not the initial noise) using dedicated spectral encoders/decoders end-to-end, with a cross-band consistency constraint enforcing aligned identity signals across complementary semantic and textural spectral regions. Compared to WMAdapter~\cite{ci2024wmadapteraddingwatermarkcontrol} and Fourier Integrity~\cite{lee2025semanticwatermarkingreinventedenhancing}, BiSLW achieves its robustness–fidelity trade-off without adapter fine-tuning or backbone changes.

\section{Method}

\subsection{Overview}

\begin{figure}[t]
    \centering
    \includegraphics[width=\linewidth]{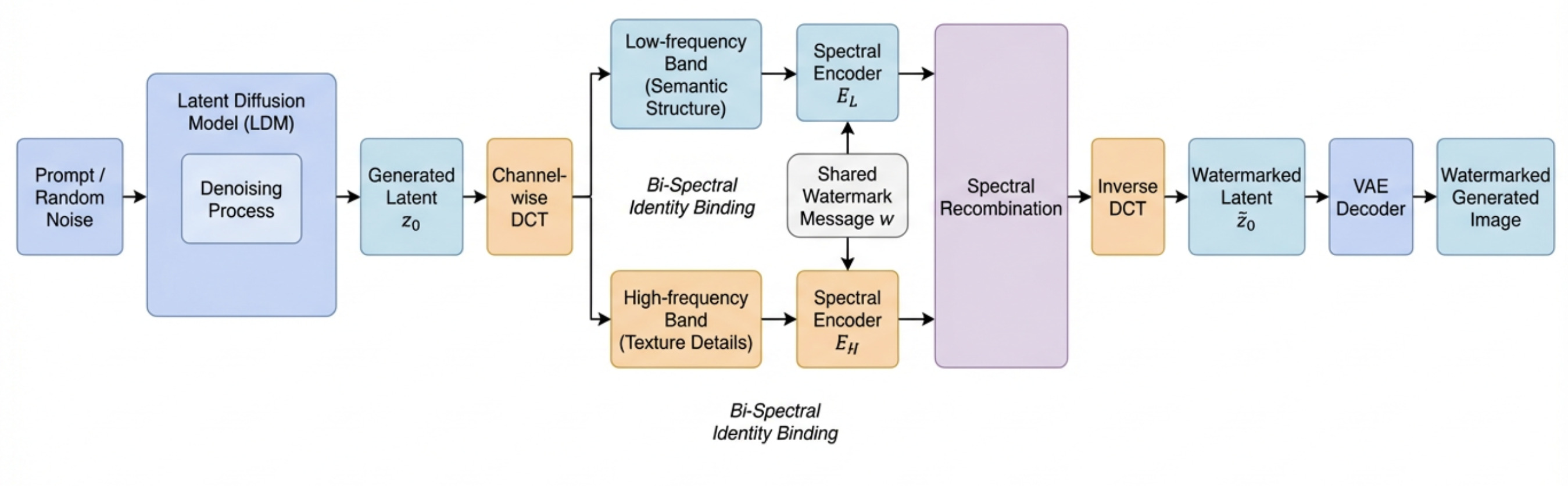}
    \caption{
        BiSLW pipeline overview. The generated latent is spectrally decomposed via channel-wise DCT into semantic (low-frequency) and textural (high-frequency) bands. A shared watermark message is independently embedded into both bands before inverse transformation and decoding.
    }
    \label{fig:pipeline}
\end{figure}

We introduce \textbf{BiSLW} (\textbf{Bi}-\textbf{S}pectral \textbf{L}atent \textbf{W}atermarking), a framework that embeds identity signals directly within the generative process of latent diffusion models (LDMs)~\cite{rombach2022highresolutionimagesynthesislatent}.  
Unlike conventional approaches that modify the pixel domain after image synthesis~\cite{zhu2018hiddenhidingdatadeep,fernandez2023stablesignaturerootingwatermarks}, BiSLW operates entirely in latent space, allowing watermark information to become an intrinsic property of the generated image.

Let $\mathbf{w} \in \{-1,+1\}^{d_w}$ denote a binary watermark message and $\mathbf{z}_0 \in \mathbb{R}^{C \times H \times W}$ the clean latent produced by the reverse diffusion process. Our objective is to construct a watermarked latent $\tilde{\mathbf{z}}_0$ such that:

\begin{itemize}
\item the decoded image $\tilde{\mathbf{x}} = \mathcal{D}(\tilde{\mathbf{z}}_0)$ remains perceptually indistinguishable from $\mathbf{x} = \mathcal{D}(\mathbf{z}_0)$,
\item the watermark $\mathbf{w}$ can be dependably extracted subsequent to distortions or regeneration attacks,
\end{itemize}

We observe that diffusion latent representations naturally contain hierarchical frequency components: low-frequency components encode global semantic structure, while high-frequency components capture fine-grained texture. BiSLW exploits this structure by independently embedding the watermark into both spectral bands, establishing complementary redundancy that strengthens robustness across a wide range of attack scenarios. The complete pipeline is illustrated in Fig.~\ref{fig:pipeline}.

% ----------------------------------------------------------
\subsection{Latent Spectral Decomposition}
\label{sec:spectral}
% ----------------------------------------------------------

\begin{figure}[t]
    \centering
    \includegraphics[width=\linewidth]{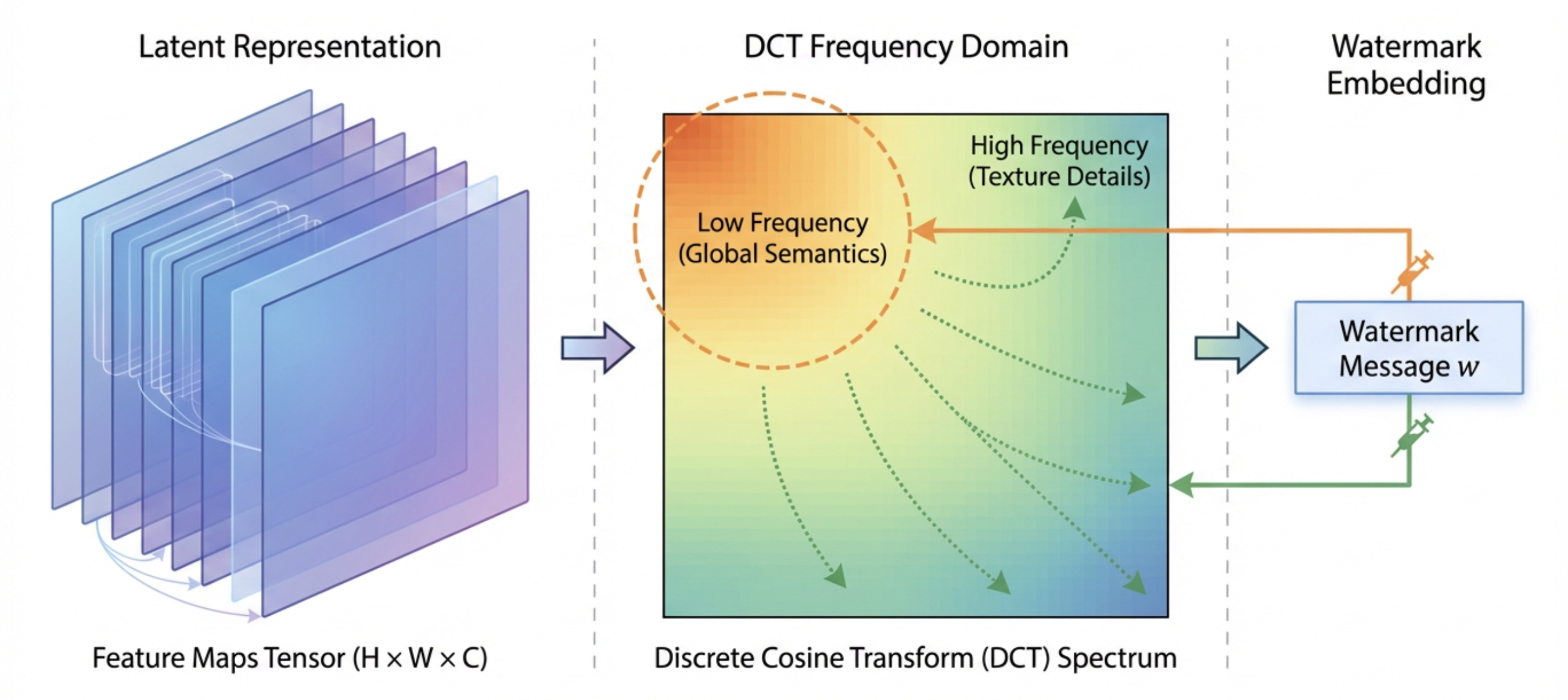}
    \caption{
        Frequency-domain interpretation of diffusion latents. Applying a channel-wise DCT to the latent tensor exposes two complementary components: low-frequency coefficients capturing coarse semantic layout and high-frequency coefficients encoding local texture. BiSLW targets both regions for watermark embedding, creating structured spectral redundancy.
    }
    \label{fig:spectral}
\end{figure}

Latent diffusion models represent images in a compressed latent space obtained
through a variational autoencoder (VAE) with encoder $\mathcal{E}$ and decoder
$\mathcal{D}$~\cite{kingma2022autoencodingvariationalbayes}. The latent tensor $\mathbf{z}_0$ produced
after the reverse diffusion process exhibits structured spatial variation
suitable for frequency analysis, as illustrated in Fig.~\ref{fig:spectral}.

We transform the latent representation into the frequency domain using a
channel-wise two-dimensional discrete cosine transform (DCT):
\begin{equation}
\mathbf{Z}^{\mathrm{freq}}_c =
\mathrm{DCT}_{2D}(\mathbf{z}_{0,c}), \quad c = 1,\ldots,C .
\label{eq:dct}
\end{equation}

Stacking all channels yields the spectral tensor

\begin{equation}
\mathbf{Z}^{\mathrm{freq}} =
\mathcal{T}(\mathbf{z}_0),
\label{eq:dct_stack}
\end{equation}

where $\mathcal{T}$ denotes the channel-wise DCT operator.

Because most signal energy concentrates in low-frequency coefficients~\cite{Rao1990DiscreteCT}, we partition the spectrum using a radial binary mask $\mathcal{M}$:

\begin{equation}
\mathcal{M}[i,j] =
\begin{cases}
1 & \text{if } \sqrt{(i/H)^2 + (j/W)^2} \le r \\
0 & \text{otherwise}
\end{cases}
\label{eq:mask}
\end{equation}

Prior to splitting, we apply channel-wise $z$-score normalisation to $\mathbf{Z}^{\mathrm{freq}}$ to account for the orders-of-magnitude difference between the DC coefficient and higher-frequency AC coefficients, which stabilises encoder network training. The normalisation is inverted exactly before the inverse DCT step.

This produces two complementary components
\begin{align}
\mathbf{Z}^{\mathrm{low}} &= \mathbf{Z}^{\mathrm{freq}} \odot \mathcal{M} \\
\mathbf{Z}^{\mathrm{high}} &= \mathbf{Z}^{\mathrm{freq}} \odot (1-\mathcal{M})
\end{align}

which capture semantic layout and fine texture respectively. The decomposition is lossless since
\begin{align}
\mathbf{Z}^{\mathrm{low}} + \mathbf{Z}^{\mathrm{high}} = \mathbf{Z}^{\mathrm{freq}} .
\end{align}
% ----------------------------------------------------------
\subsection{Bi-Spectral Watermark Embedding}
\label{sec:embedding}
% ----------------------------------------------------------

We embed the watermark into both spectral bands using two lightweight perturbation networks $\Delta_L$ and $\Delta_H$. Each network predicts a residual conditioned on the watermark message $\mathbf{w}$.

The modified spectral bands are
\begin{align}
\tilde{\mathbf{Z}}^{\mathrm{low}} &= \mathbf{Z}^{\mathrm{low}} + \alpha_L \Delta_L(\mathbf{Z}^{\mathrm{low}},\mathbf{w}) \\
\tilde{\mathbf{Z}}^{\mathrm{high}} &= \mathbf{Z}^{\mathrm{high}} + \alpha_H \Delta_H(\mathbf{Z}^{\mathrm{high}},\mathbf{w})
\end{align}

where $\alpha_L$ and $\alpha_H$ control embedding strength. In practice we set $\alpha_L > \alpha_H$ because low-frequency perturbations tend to survive stronger distortions.

The watermarked spectrum is reconstructed as
\begin{equation}
\tilde{\mathbf{Z}}^{\mathrm{freq}} =
\tilde{\mathbf{Z}}^{\mathrm{low}} + \tilde{\mathbf{Z}}^{\mathrm{high}}
\end{equation}

and the final latent is obtained via inverse DCT:

\begin{equation}
\tilde{\mathbf{z}}_0 = \mathcal{T}^{-1}(\tilde{\mathbf{Z}}^{\mathrm{freq}})
\end{equation}

Crucially, this embedding takes place prior to VAE decoding, ensuring the watermark remains intrinsic to the generative representation.

\textbf{Network architecture.}
Both perturbation networks share a common architectural blueprint: a lightweight convolutional encoder–decoder backbone with residual connections~\cite{he2015deepresiduallearningimage}. Conditioning on the watermark message is achieved through Feature-wise Linear Modulation (FiLM)~\cite{perez2017filmvisualreasoninggeneral}, which maps $\mathbf{w}$ to per-channel scaling and bias parameters.

% ----------------------------------------------------------
\subsection{Watermark Extraction}
\label{sec:extraction}
% ----------------------------------------------------------

Given a possibly distorted image $\tilde{\mathbf{x}}'$, watermark extraction proceeds as follows:

\begin{enumerate}
\item Encode image to latent space: $\mathbf{z}'=\mathcal{E}(\tilde{\mathbf{x}}')$
\item Apply channel-wise DCT: $\mathbf{Z}'^{\mathrm{freq}}=\mathcal{T}(\mathbf{z}')$
\item Split spectral components using mask $\mathcal{M}$
\item Decode watermark from each band
\end{enumerate}
\begin{equation}
\hat{\mathbf{w}}_L = D_L(\mathbf{Z}'^{\mathrm{low}})
\qquad
\hat{\mathbf{w}}_H = D_H(\mathbf{Z}'^{\mathrm{high}})
\end{equation}

The final estimate is obtained by averaging
\begin{equation}
\hat{\mathbf{w}} = \frac{1}{2}(\hat{\mathbf{w}}_L+\hat{\mathbf{w}}_H)
\end{equation}
followed by a sign operation to recover the binary message.

% ----------------------------------------------------------
\subsection{Cross-Band Identity Consistency}
\label{sec:consistency}
% ----------------------------------------------------------

Embedding the watermark in two spectral domains introduces the risk of spectral drift, where the two decoders predict inconsistent identities. To enforce alignment we introduce a cross-band consistency loss

\begin{equation}
\mathcal{L}_{cons} =
\left\|\hat{\mathbf{w}}_L - \hat{\mathbf{w}}_H\right\|_2^2 .
\end{equation}

This constraint encourages both spectral pathways to encode the same identity signal, creating redundancy that improves robustness under distortions.

% ----------------------------------------------------------
\subsection{Training Objective}
\label{sec:loss}
% ----------------------------------------------------------

\begin{figure*}[t]
    \centering
    \includegraphics[width=\textwidth]{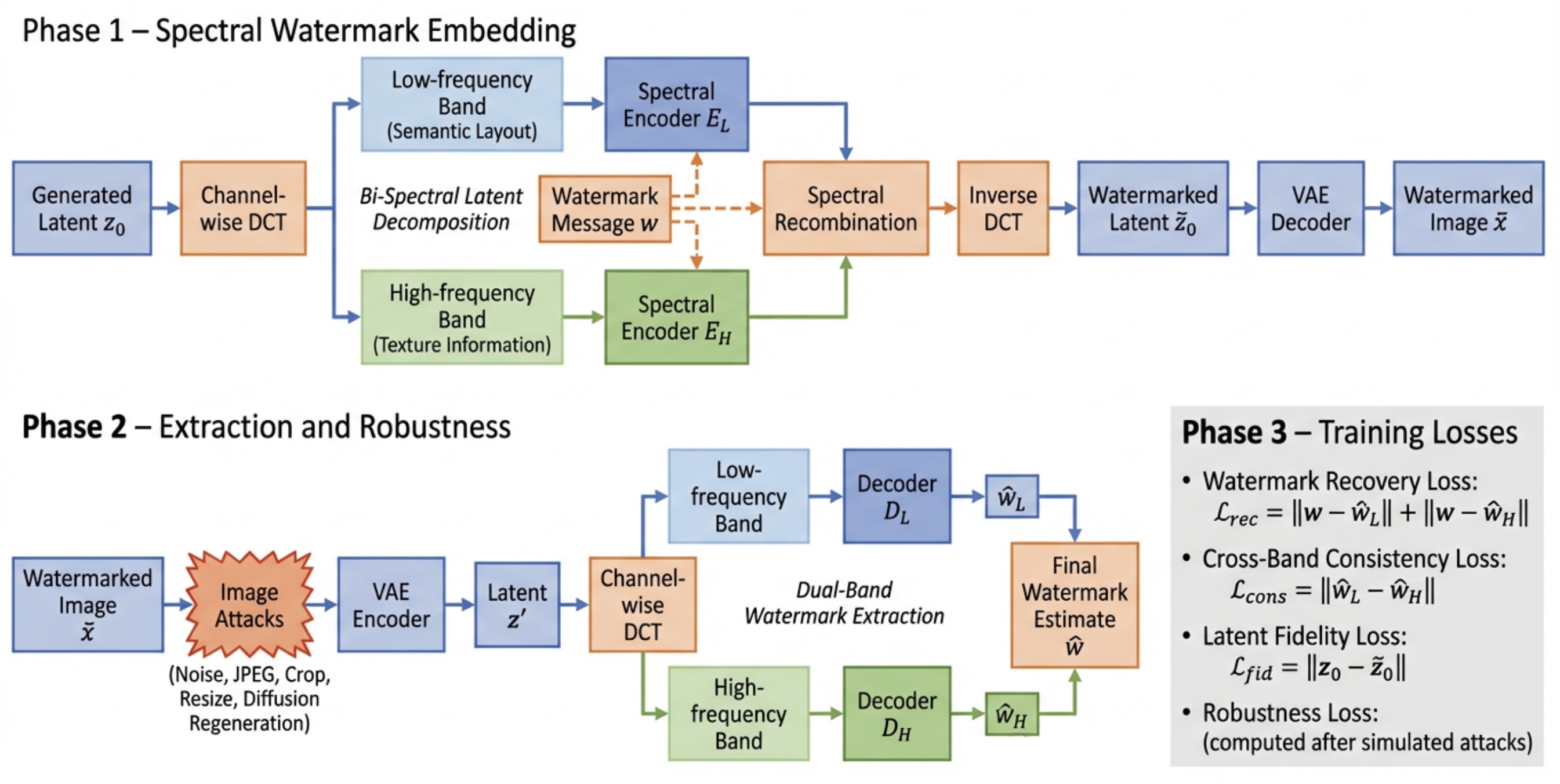}
    \caption{
        BiSLW training framework. Spectral watermark embedding and attack-conditioned extraction are jointly optimized using a combination of watermark recovery, cross-band consistency, and latent fidelity losses.
    }
    \label{fig:framework}
\end{figure*}

We jointly train the embedding networks $\Delta_L,\Delta_H$ and extraction networks $D_L,D_H$ using a multi-objective loss.

{Watermark recovery.}

\begin{equation}
\mathcal{L}_w =
\|\hat{\mathbf{w}}_L-\mathbf{w}\|_2^2 +
\|\hat{\mathbf{w}}_H-\mathbf{w}\|_2^2
\end{equation}

{Latent fidelity.}

\begin{equation}
\mathcal{L}_z =
\|\tilde{\mathbf{z}}_0-\mathbf{z}_0\|_2^2
\end{equation}

{Robustness loss.}

\begin{equation}
\mathcal{L}_{rob} =
\mathbb{E}_{\mathcal{A}\sim\mathcal{P_A}}
\left[
\|\hat{\mathbf{w}}_L^{\mathcal{A}}-\mathbf{w}\|_2^2 +
\|\hat{\mathbf{w}}_H^{\mathcal{A}}-\mathbf{w}\|_2^2
\right]
\end{equation}

Here $\mathcal{P_A}$ denotes the distribution over the differentiable
attack pool described in Sec.~\ref{sec:robustness} (JPEG compression,
Gaussian blur, noise, random crop, and diffusion regeneration), and the
expectation is approximated by sampling one attack per training step.

The final training objective is
\begin{equation}
\mathcal{L} =
\lambda_w\mathcal{L}_w +
\lambda_{cons}\mathcal{L}_{cons} +
\lambda_z\mathcal{L}_z +
\lambda_{rob}\mathcal{L}_{rob}.
\end{equation}

The diffusion backbone remains frozen during training; only the watermark embedding and extraction networks are optimized. The full training pipeline is illustrated in Fig.~\ref{fig:framework}.

% ----------------------------------------------------------
\subsection{Regeneration Robustness}
\label{sec:robustness}
% ----------------------------------------------------------

Regeneration attacks re-sample images through the diffusion process to remove embedded signals. Formally,

\begin{equation}
\mathcal{A}_{regen}(\tilde{\mathbf{x}})
=
\mathcal{D}\big(
f_\theta(\sqrt{\bar{\alpha}_{t^*}}\mathcal{E}(\tilde{\mathbf{x}})
+\sqrt{1-\bar{\alpha}_{t^*}}\epsilon ,t^*)
\big)
\end{equation}

where $\epsilon\sim\mathcal{N}(0,I)$ and $t^*$ denotes the regeneration timestep.

During training we simulate regeneration attacks together with common image distortions including JPEG compression, Gaussian blur, noise, and random cropping. These augmentations are incorporated through the robustness loss to improve watermark recovery under realistic conditions.

% ----------------------------------------------------------
\subsection{Discussion}
\label{sec:discussion}
% ----------------------------------------------------------

\textbf{Comparison to spatial latent watermarking.}
Existing latent watermarking methods embed their signals directly into the spatial latent representation~\cite{wen2023treeringwatermarksfingerprintsdiffusion, fernandez2023stablesignaturerootingwatermarks}, concentrating the entire watermark within a single representational mode. This makes such approaches inherently vulnerable when the representation is disturbed through regeneration, the watermark signal has no fallback. BiSLW distributes identity information across complementary spectral bands, providing redundancy that improves resilience to structural perturbations.

\textbf{Quality–robustness balance.}
Embedding strengths $\alpha_L$ and $\alpha_H$ serve as tunable parameters that govern the trade-off between perceptual quality and watermark robustness. Adjusting these parameters allows practitioners to balance imperceptibility and resilience depending on application requirements.

\textbf{Empirical validation of the frequency hierarchy.}
To validate our core assumption, we performed a perturbation analysis in latent frequency space: perturbing low-frequency DCT coefficients affects coarse scene layout, while perturbing high-frequency coefficients alters local texture while preserving overall composition. Table~\ref{tab:spectral_ablation} further supports this: low-frequency-only embedding achieves higher PSNR (38.96\,dB) while high-frequency-only embedding yields better combined-attack robustness (0.90). Extended visualisations are in the supplementary material.

\section{Experiments and Results}
\label{sec:experiments}

% ----------------------------------------------------------
\subsection{Experimental Settings}
\label{sec:settings}
% ----------------------------------------------------------

\textbf{Dataset and Preprocessing.}
BiSLW's watermark embedding and extraction modules are trained on 100,000 images sourced from the MIRFLICKR-1M dataset~\cite{Huiskes2008TheMF}, a large-scale and visually diverse collection of real-world photographs gathered from Flickr under Creative Commons licenses. Prior to training, each image is randomly cropped to a fixed resolution of $256 \times 256$ pixels, following standard practice in latent diffusion watermarking~\cite{fernandez2023stablesignaturerootingwatermarks}. No additional data augmentation is applied at the dataset level, as the
differentiable attack pool described in Sec.~\ref{sec:robustness} serves as the primary source of training-time diversity.

\textbf{Baselines.}
We compare BiSLW against representative watermarking methods spanning
classical frequency-domain and modern neural approaches:
DCT-DWT~\cite{Yldz2023DigitalIW} (classical frequency-domain embedding),
HiDDeN~\cite{zhu2018hiddenhidingdatadeep} (end-to-end neural watermarking),
SSL~\cite{fernandez2022watermarkingimagesselfsupervisedlatent} (self-supervised feature-space watermarking),
RoSteALS~\cite{bui2023rostealsrobuststeganographyusing} (latent diffusion sampling-time embedding),
RivaGAN~\cite{zhang2019robustinvisiblevideowatermarking} (GAN-based robust watermarking),
FNNS~\cite{sym17071094} (fixed neural network steganography),
Stable Signature (SS)~\cite{fernandez2023stablesignaturerootingwatermarks} (decoder fine-tuning for LDMs),
and Tree-Ring~\cite{wen2023treeringwatermarksfingerprintsdiffusion}
(frequency-domain generation-time watermarking).
We also compare against LaWa~\cite{rezaei2025lawausinglatentspace} in both its in-generation ($\ast$)
and post-generation variants, which are the most directly comparable
latent watermarking baselines.
WMAdapter~\cite{ci2024wmadapteraddingwatermarkcontrol} and Fourier Integrity~\cite{lee2025semanticwatermarkingreinventedenhancing}
were identified during the review process; as full retraining was not
feasible within the revision timeline, we include a qualitative
discussion of their relationship to BiSLW in Sec.~\ref{sec:related}.
All other baselines are evaluated using their official implementations and
recommended hyperparameters.

\textbf{Evaluation metrics.}
Perceptual image quality is assessed using Peak Signal-to-Noise Ratio (PSNR, in dB) and the Structural Similarity Index (SSIM)~\cite{Wang2004ImageQA}, both computed by comparing the watermarked image against its un-watermarked counterpart. To evaluate generative fidelity beyond pixel-level metrics, we additionally report FID, CLIP similarity, and KL divergence shift (Table~\ref{tab:generative}). The extraction performance of each method is assessed using \emph{bit accuracy}, computed as the mean fraction of correctly recovered message bits over the entire evaluation set. Embedding time is reported in milliseconds per image on a single NVIDIA A100 GPU.

\textbf{Attack protocol.}
Following the standard evaluation protocol of~\cite{fernandez2023stablesignaturerootingwatermarks,fernandez2022watermarkingimagesselfsupervisedlatent},
we assess robustness under nine individual distortions and one combined attack:
center crop ($10\%$ removed), random crop ($10\%$ removed), resize ($0.7\times$),
rotation ($15^\circ$), Gaussian blur, contrast adjustment ($\times 2.0$),
brightness adjustment ($\times 2.0$), JPEG compression (quality $70$), and
a combined attack that applies all distortions sequentially.

\textbf{Implementation details.}
The perturbation encoders $\Delta_L$, $\Delta_H$ and decoders $D_L$, $D_H$ are 
trained for 40 epochs using the AdamW optimizer~\cite{loshchilov2019decoupledweightdecayregularization} with a learning rate of $6 \times 10^{-5}$ and batch size 8. Loss weights are set to $\lambda_w = 1.0$, $\lambda_{\mathrm{cons}} = 0.5$, $\lambda_{\mathrm{rob}} = 1.0$, and $\lambda_z = 2.0$ (see ablation in Sec.~\ref{sec:ablation}). The spectral split radius is set to $r = 0.25$. Embedding strength scalars are fixed at  $\alpha_L = 0.8$ and $\alpha_H = 0.3$. The backbone LDM is frozen throughout. HiDDeN is evaluated using its publicly available 30-bit checkpoint; all other methods are evaluated at the bit capacity reported in Table~\ref{tab:main}. Code and pre-trained models are publicly available at \url{https://github.com/OVER-CODER/BiSLW}.

% ----------------------------------------------------------
\subsection{Comparison Results}
\label{sec:comparison}
% ----------------------------------------------------------

Table~\ref{tab:main} reports results on AI-generated images (32-bit watermark) and CLIC dataset (48-bit watermark).

\textbf{Image quality.}
BiSLW achieves a PSNR of 37.40\,dB / SSIM of 0.91 on AI-generated images.
BiSLW provides a favorable balance between perceptual fidelity and robustness, outperforming prior latent diffusion watermarking methods at the same bit capacity.
Notably, BiSLW exceeds LaWa — the next strongest method in terms of PSNR —
by approximately 3\,dB while simultaneously matching it on robustness.
On CLIC (48-bit), BiSLW likewise improves upon the best prior method (FNNS,
36.84\,dB) while providing far superior robustness under the combined attack
(0.94 vs.\ 0.91).

\textbf{Robustness.}
BiSLW achieves near-perfect bit accuracy under most distortions on both benchmarks, matching LaWa while surpassing all other
baselines on the challenging combined attack (0.98 on AI-generated, 0.94 on CLIC).
Classical methods such as DCT-DWT collapse to near-chance accuracy ($\approx 0.51$)
under most geometric distortions, while neural baselines such as HiDDeN suffer
on JPEG compression (0.53) and combined attacks (0.59).
SSL achieves strong robustness but at an embedding time of 870\,ms, nearly
three orders of magnitude slower than BiSLW (1\,ms).

\textbf{Embedding efficiency.}
The in-generation variant of BiSLW ($\ast$) incurs only 1\,ms overhead per image — equivalent to LaWa and Stable Signature — because the four lightweight networks operate on the compact latent tensor rather than in pixel space. The post-generation variant adds 33\,ms, comparable to HiDDeN but with substantially better fidelity and robustness.

\textbf{Generative quality and latent preservation.}
Table~\ref{tab:generative} reports FID, CLIP similarity, KL divergence shift, and regeneration robustness for key methods. BiSLW achieves FID of 9.0, closer to vanilla Stable Diffusion (8.7) than either LaWa (9.8) or Tree-Ring (9.9), confirming that the watermark embedding does not measurably degrade the generative distribution. The KL shift of 0.018 and latent shift of 0.011 are the lowest among watermarking methods, suggesting that BiSLW best preserves the diffusion latent manifold despite embedding perturbations. At regeneration timestep $t^*\!=\!250$, BiSLW maintains 0.96 bit accuracy vs.\ 0.94 for LaWa and 0.88 for Tree-Ring, further demonstrating the benefit of the bi-spectral design.

\begin{table*}[t]
\centering
\caption{Comparison with existing watermarking methods on AI-generated 
images and the CLIC dataset. $\uparrow$/$\downarrow$ denote higher/lower is better.}
\label{tab:main}
\resizebox{\textwidth}{!}{
\begin{tabular}{l|c|c|cccccccccc|c}
\toprule
& \multicolumn{1}{c|}{\textbf{Image quality}} 
& \multirow{2}{*}{\makecell{\textbf{Emb.}\\\textbf{time}\\(ms)}} 
& \multicolumn{10}{c|}{\textbf{Bit accuracy $\uparrow$}} 
& \multirow{2}{*}{\textbf{Avg.}} \\
\textbf{Method (bit\#)} 
& \textbf{PSNR/SSIM $\uparrow$}
&
& \rotatebox{60}{None}
& \rotatebox{60}{C. Crop 0.1}
& \rotatebox{60}{R. Crop 0.1}
& \rotatebox{60}{Resize 0.7}
& \rotatebox{60}{Rot. 15}
& \rotatebox{60}{Blur}
& \rotatebox{60}{Contr. 2.0}
& \rotatebox{60}{Bright. 2.0}
& \rotatebox{60}{JPEG 70}
& \rotatebox{60}{Comb.}
&  \\
\midrule
\multicolumn{14}{c}{\textbf{AI-generated Images}} \\
\midrule
DCT-DWT (32) & 39.47/0.97 & 139 & 0.91 & 0.51 & 0.51 & 0.52 & 0.52 & 0.51 & 0.52 & 0.51 & 0.51 & 0.51 & 0.55 \\
HiDDeN (30) & 32.59/0.95 & 17 & 0.91 & 0.91 & 0.91 & 0.82 & 0.79 & 0.76 & 0.75 & 0.74 & 0.53 & 0.59 & 0.77 \\
SSL (32) & 33.23/0.89 & 870 & 1.00 & 0.74 & 0.72 & 0.99 & 0.99 & 1.00 & 0.96 & 0.95 & 0.99 & 0.85 & 0.92 \\
RoSteALS (32) & 29.31/0.93 & 144 & 1.00 & 0.50 & 0.51 & 1.00 & 0.47 & 1.00 & 0.88 & 0.86 & 1.00 & 0.50 & 0.77 \\
RivaGAN (32) & 40.53/0.98 & 57 & 0.99 & 0.98 & 0.98 & 0.87 & 0.91 & 0.99 & 0.81 & 0.80 & 0.98 & 0.93 & 0.92 \\
Tree-Ring (32) & 33.64/0.90 & 46 & 0.88 & 0.81 & 0.79 & 0.85 & 0.72 & 0.88 & 0.87 & 0.86 & 0.83 & 0.81 & 0.83 \\
\midrule
LaWa$^\ast$ (32) & 34.25/0.89 & 1 & 1.00 & 1.00 & 0.98 & 1.00 & 1.00 & 1.00 & 1.00 & 1.00 & 1.00 & 0.97 & 0.99 \\
LaWa-post-gen (32) & 34.28/0.90 & 33 & 1.00 & 1.00 & 0.96 & 1.00 & 1.00 & 1.00 & 1.00 & 1.00 & 1.00 & 0.98 & 0.99 \\
\midrule
\textbf{BiSLW$^\ast$ (32)} & \underline{37.40}/\underline{0.91} & \textbf{1} & \textbf{1.00} & \textbf{1.00} & \textbf{0.99} & \textbf{1.00} & \textbf{1.00} & \textbf{1.00} & \textbf{1.00} & \textbf{1.00} & \textbf{1.00} & \underline{0.98} & \textbf{0.99} \\
\textbf{BiSLW-post-gen (32)} & \textbf{37.42}/\textbf{0.91} & 33 & \textbf{1.00} & \textbf{1.00} & 0.97 & \textbf{1.00} & \textbf{1.00} & \textbf{1.00} & \textbf{1.00} & \textbf{1.00} & \textbf{1.00} & \textbf{0.98} & \textbf{0.99} \\
\midrule
\multicolumn{14}{c}{\textbf{CLIC}} \\
\midrule
SSL (48) & 33.25/0.89 & 870 & 1.00 & 0.72 & 0.70 & 0.99 & 0.99 & 1.00 & 0.94 & 0.94 & 0.99 & 0.82 & 0.91 \\
FNNS (48) & 36.84/0.97 & 645 & 1.00 & 0.98 & 0.96 & 0.95 & 0.75 & 0.54 & 0.86 & 0.85 & 0.94 & 0.91 & 0.87 \\
RoSteALS (48) & 29.30/0.92 & 144 & 1.00 & 0.51 & 0.50 & 1.00 & 0.50 & 1.00 & 0.88 & 0.85 & 1.00 & 0.49 & 0.78 \\
SS$^\ast$ (48) & 32.02/0.84 & 0 & 0.99 & 0.95 & 0.93 & 0.96 & 0.81 & 0.78 & 0.97 & 0.96 & 0.92 & 0.92 & 0.92 \\
\midrule
LaWa$^\ast$ (48) & 33.52/0.86 & 1 & 1.00 & 0.95 & 0.91 & \underline{0.99} & 0.96 & 0.99 & 1.00 & 1.00 & 1.00 & 0.94 & 0.97 \\
LaWa-post-gen (48) & 33.45/0.87 & 34 & 1.00 & 0.94 & 0.90 & 0.96 & 0.97 & 0.99 & 1.00 & 1.00 & 1.00 & 0.95 & 0.97 \\
\midrule
\textbf{BiSLW$^\ast$ (48)} & \underline{36.82}/\underline{0.90} & \textbf{1} & \textbf{1.00} & \underline{0.96} & 0.90 & 0.98 & \underline{0.97} & \textbf{0.99} & \textbf{1.00} & \textbf{1.00} & \textbf{1.00} & \underline{0.95} & \textbf{0.98} \\
\textbf{BiSLW-post-gen (48)} & \textbf{37.10}/\textbf{0.91} & 34 & \textbf{1.00} & \underline{0.95} & 0.90 & \underline{0.97} & \textbf{0.98} & \textbf{0.99} & \textbf{1.00} & \textbf{1.00} & \textbf{1.00} & \textbf{0.96} & \textbf{0.98} \\
\bottomrule
\end{tabular}
}
\end{table*}

\begin{table}[t]
\centering
\caption{Generative quality and latent preservation comparison.
FID and KL Shift are lower-is-better; CLIP and Regen.\ accuracy are
higher-is-better. Regen.\ reports bit accuracy at $t^*\!=\!250/500$.}
\label{tab:generative}
\small
\setlength{\tabcolsep}{4pt}
\resizebox{\textwidth}{!}{
\begin{tabular}{lcccccc}
\toprule
\textbf{Method} & \textbf{FID$\downarrow$} & \textbf{CLIP$\uparrow$} & \textbf{PSNR$\uparrow$} & \textbf{KL Shift$\downarrow$} & \textbf{Lat.\ Shift$\downarrow$} & \textbf{Regen.\ (250/500)$\uparrow$} \\
\midrule
Vanilla SD      & 8.7 & 0.312 & —     & —     & —     & — \\
LaWa$^\ast$~\cite{rezaei2025lawausinglatentspace}
                & 9.8 & 0.307 & 34.25 & 0.021 & 0.014 & 0.94 / 0.89 \\
Tree-Ring~\cite{wen2023treeringwatermarksfingerprintsdiffusion}
                & 9.9 & 0.302 & 33.64 & 0.024 & 0.018 & 0.88 / 0.81 \\
\textbf{BiSLW$^\ast$ (Ours)}
                & \textbf{9.0} & \textbf{0.311} & \textbf{37.40} & \textbf{0.018} & \textbf{0.011} & \textbf{0.96 / 0.92} \\
\bottomrule
\end{tabular}
}
\end{table}

\begin{figure}[t]
\centering
\includegraphics[width=0.8\linewidth]{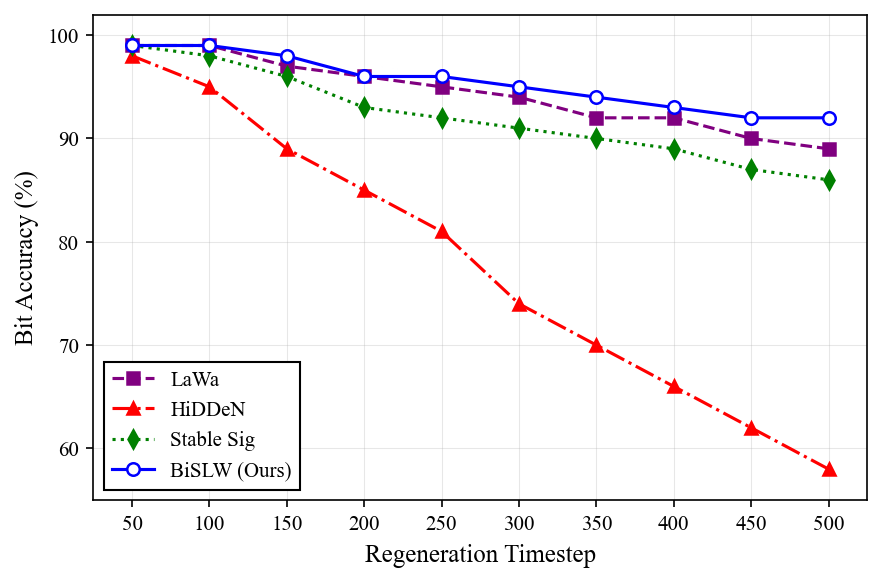}
\caption{
Robustness to diffusion regeneration attacks under increasing regeneration timestep $t^*$.
Increasing the regeneration timestep raises the level of noise injected into the latent representation before denoising begins.
BiSLW maintains significantly higher bit accuracy compared to prior watermarking
methods, demonstrating the benefit of bi-spectral embedding across both semantic
(low-frequency) and textural (high-frequency) latent components.
}
\label{fig:regen}
\end{figure}

\subsection{Robustness to Diffusion Regeneration}

Diffusion regeneration is a challenging attack scenario in which an
image is re-sampled through the diffusion process, potentially removing
embedded signals. To evaluate robustness under this setting, we vary the
regeneration timestep $t^*$ from 50 to 500, where larger values correspond
to stronger noise injection before denoising.

Fig.~\ref{fig:regen} compares the bit extraction accuracy of BiSLW with
representative watermarking baselines under increasing regeneration strength.
BiSLW consistently maintains the highest decoding accuracy across all
timesteps. In particular, accuracy remains above 96\% up to $t^* = 250$
and gradually decreases to 92\% at $t^* = 500$.

In contrast, pixel-domain watermarking methods such as HiDDeN exhibit
severe degradation as regeneration strength increases, while methods
designed for generative models (e.g., Stable Signature and LaWa)
show moderate performance drops. These results confirm that embedding
the watermark across both spectral components of the latent space
provides stronger resilience to regeneration attacks.

% ----------------------------------------------------------
\subsection{Qualitative Analysis}
\label{sec:qualitative}
% ----------------------------------------------------------

\begin{figure*}[t]
\centering
\setlength{\tabcolsep}{2pt}
\renewcommand{\arraystretch}{1.1}

\begin{tabular}{c c c c c}

\rotatebox{90}{\shortstack{{Original}\\{Generation}}} &
\includegraphics[width=0.19\textwidth]{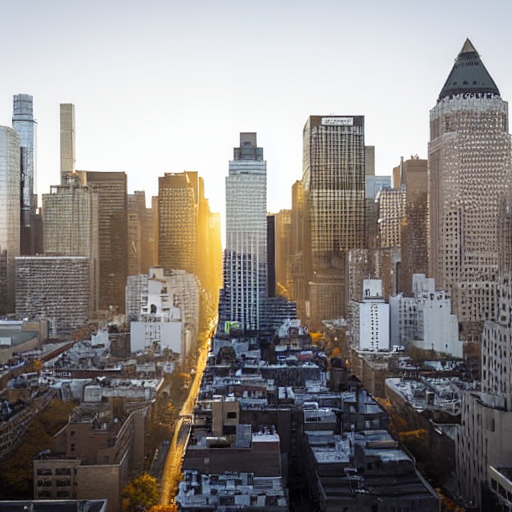} &
\includegraphics[width=0.19\textwidth]{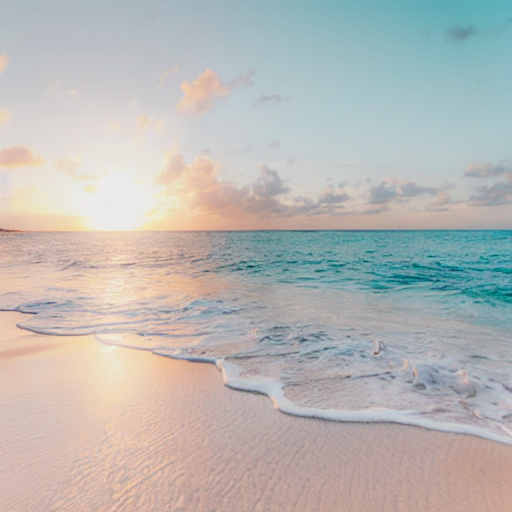} &
\includegraphics[width=0.19\textwidth]{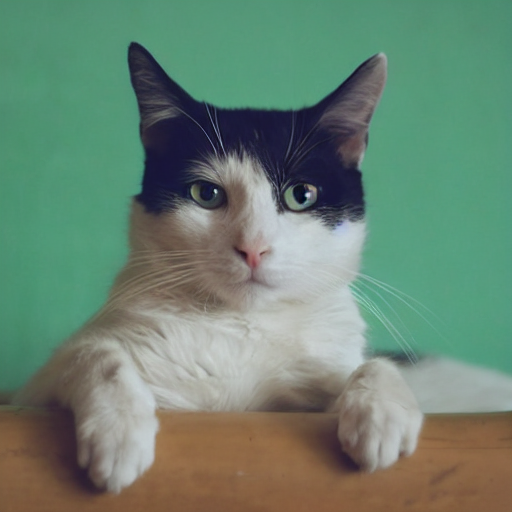} &
\includegraphics[width=0.19\textwidth]{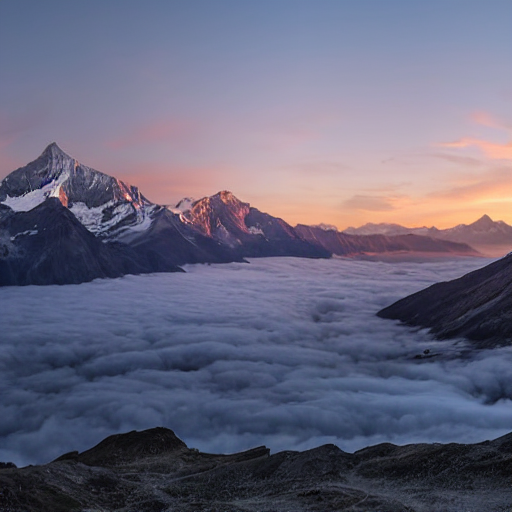} \\

\rotatebox{90}{\shortstack{{BiSLW}\\{Generation}}} &
\includegraphics[width=0.19\textwidth]{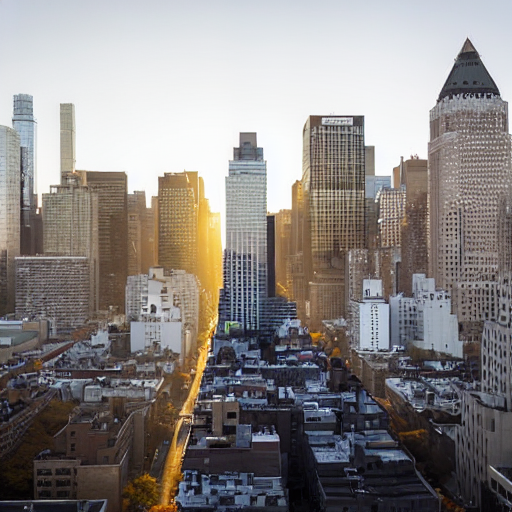} &
\includegraphics[width=0.19\textwidth]{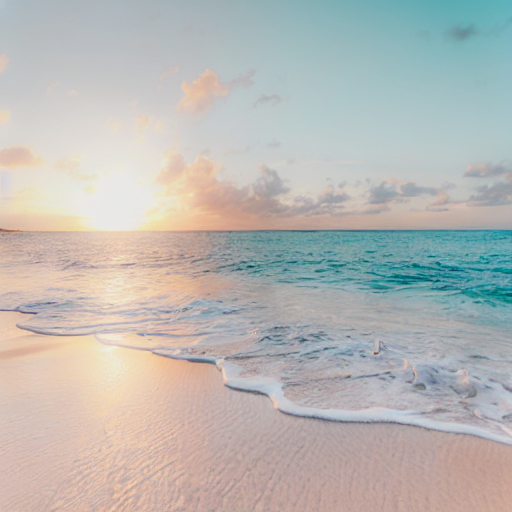} &
\includegraphics[width=0.19\textwidth]{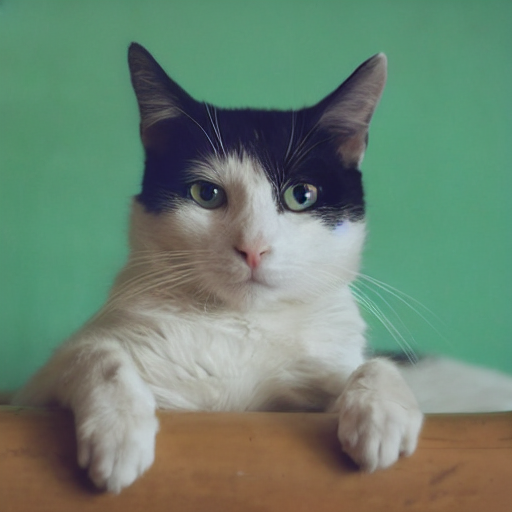} &
\includegraphics[width=0.19\textwidth]{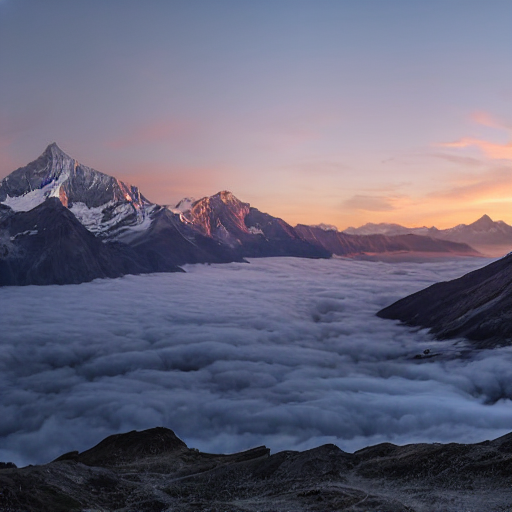} \\

\rotatebox{90}{{Residual }$\times 10$} &
\includegraphics[width=0.19\textwidth]{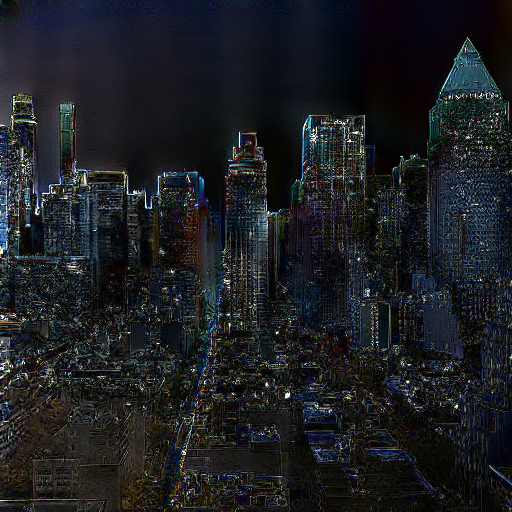} &
\includegraphics[width=0.19\textwidth]{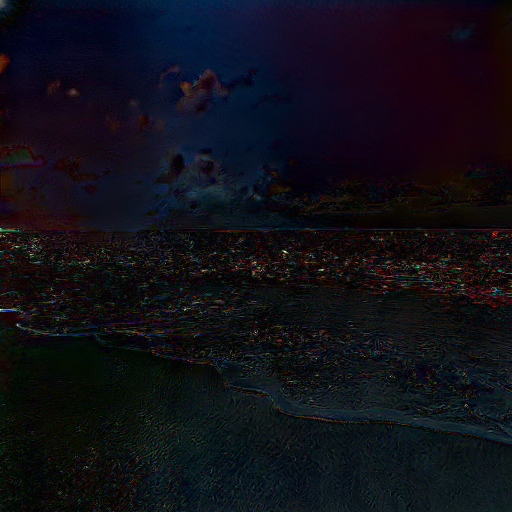} &
\includegraphics[width=0.19\textwidth]{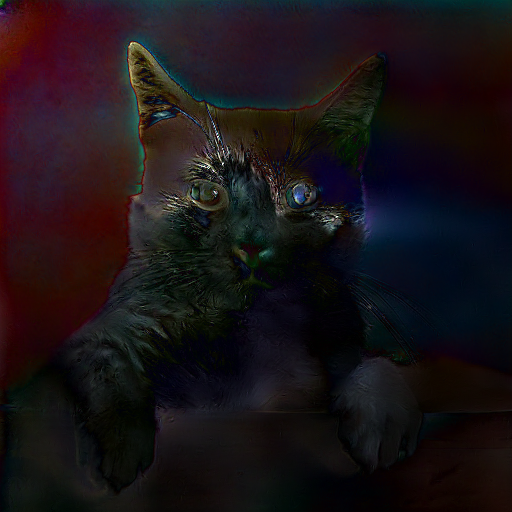} &
\includegraphics[width=0.19\textwidth]{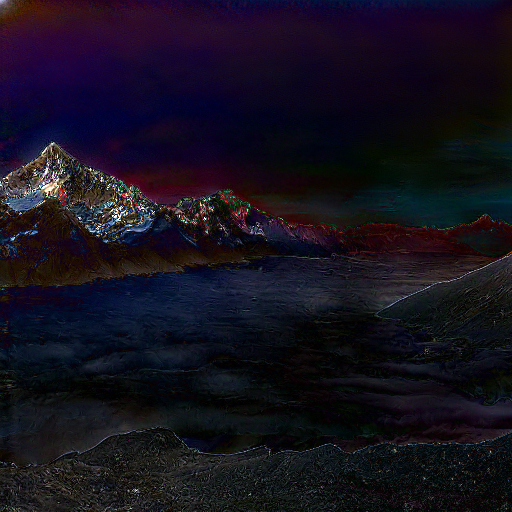} \\

\shortstack{PSNR/\\SSIM} &
\textbf{40.24 / 0.98} &
\textbf{38.35 / 0.96} &
\textbf{39.11 / 0.96} &
\textbf{36.74 / 0.93} \\

\end{tabular}

\caption{
Qualitative comparison between original generations and images watermarked with \textbf{BiSLW}. 
Residual maps (×10) highlight watermark perturbations while preserving visual fidelity. Despite structured spectral embedding, BiSLW preserves high perceptual fidelity, as reflected by consistently high PSNR/SSIM values across diverse scenes.
}

\label{fig:qualitative}
\end{figure*}

Fig.~\ref{fig:qualitative} presents a visual comparison between watermarked images produced by BiSLW and their un-watermarked counterparts.
The watermarked images achieve PSNR values of 40.24\,dB, 38.35\,dB, 39.11\,dB,
and 36.74\,dB (SSIM 0.98, 0.96, 0.96, and 0.93 respectively) across four representative samples spanning different semantic categories,
confirming that the embedded perturbations remain well below the threshold of human perceptibility regardless of content type.
Unlike pixel-domain methods that can introduce subtle banding or colour shifts, particularly in smooth regions or under high embedding strength~\cite{cox2007digital}, BiSLW's latent-domain modifications are geometrically consistent with the VAE decoder's reconstruction statistics, ensuring the watermark remains naturally integrated into the image structure.

% ----------------------------------------------------------
\subsection{Quality and Robustness Trade-offs}
\label{sec:tradeoffs}
% ----------------------------------------------------------

We evaluate BiSLW's scalability by varying the embedded message length from 32 to 128 bits, reporting perceptual quality and extraction accuracy across representative distortions in Table~\ref{tab:capacity}. As payload increases, quality degrades gradually: PSNR drops from 38.82\,dB to 36.34\,dB and SSIM from 0.91 to 0.85, a modest decline given the $4\times$ capacity increase, with both metrics remaining within accepted imperceptibility bounds~\cite{Wang2004ImageQA}.

Robustness remains high at moderate capacities.
For 32 and 48 bits, BiSLW achieves perfect extraction accuracy under
most individual distortions and maintains 0.98 and 0.94 accuracy under
the combined attack respectively. At 64 bits, accuracy under center
cropping decreases slightly to 0.92 while remaining above 0.93 under
combined distortions.

At higher payloads, robustness begins to taper, most notably under rotation at 96 bits (0.77), which is expected given that rotation disrupts the high-frequency spectral band.
Notably, even at 128 bits, combined attack accuracy of 0.87 still surpasses several baselines evaluated at much smaller payloads, including HiDDeN (0.59) and RoSteALS (0.50) at just 32 bits. Overall, BiSLW sustains a strong quality–robustness balance at capacities well beyond what prior methods support.

\begin{table}[t]
\centering
\caption{
    Effect of watermark capacity on image quality and bit accuracy.
    Results reported on AI-generated images.
}
\label{tab:capacity}
\begin{tabular}{c c c c c c c c c}
\toprule
\textbf{Bits} & \textbf{PSNR (dB)$\uparrow$} & \textbf{SSIM$\uparrow$} &
\textbf{None} & \textbf{C.Crop} & \textbf{Contr.} & \textbf{Rot.} & \textbf{JPEG} & \textbf{Comb.} \\
\midrule
 32  & 38.82 & 0.91 & 1.00 & 1.00 & 1.00 & 1.00 & 1.00 & 0.98 \\
 48  & 38.47 & 0.89 & 1.00 & 0.95 & 1.00 & 0.97 & 1.00 & 0.94 \\
 64  & 37.70 & 0.87 & 0.99 & 0.92 & 0.97 & 0.96 & 0.99 & 0.93 \\
 96  & 36.88 & 0.86 & 0.99 & 0.87 & 0.96 & 0.77 & 0.94 & 0.91 \\
128  & 36.34 & 0.85 & 0.97 & 0.94 & 0.96 & 0.82 & 0.90 & 0.87 \\
\bottomrule
\end{tabular}
\end{table}

% ----------------------------------------------------------
\subsection{Ablation Study}
\label{sec:ablation}
% ----------------------------------------------------------

\begin{table}[t]
\centering
\caption{
Spectral component ablation. We isolate the contribution of 
each frequency band by comparing low-frequency-only, 
high-frequency-only, and full bi-spectral embedding variants of BiSLW 
on AI-generated images (48-bit watermark).
}
\label{tab:spectral_ablation}
\small
\setlength{\tabcolsep}{6pt}
\begin{tabular}{lccc}
\toprule
\textbf{Embedding Strategy} &
\textbf{PSNR $\uparrow$} &
\textbf{SSIM $\uparrow$} &
\textbf{Bit Acc. (Comb.) $\uparrow$} \\
\midrule
Low-frequency only  & \textbf{38.96} & \textbf{0.92} & 0.88 \\
High-frequency only & 37.12 & 0.89 & 0.90 \\
BiSLW (Low + High)  & 37.71 & 0.89 & \textbf{0.93} \\
\bottomrule
\end{tabular}
\end{table}

\begin{wraptable}{r}{0.48\linewidth}
\vspace{-10pt}
\centering
\caption{
Ablation on the latent fidelity loss weight $\lambda_z$.
Results reported on AI-generated images (48 bits).
}
\label{tab:ablation}
\scriptsize
\setlength{\tabcolsep}{3pt}
\begin{tabular}{lcccccc}
\toprule
$\lambda_z$ & 0.5 & 2.0 & 5.0 & 10.0 & 20.0 \\
\midrule
PSNR$\uparrow$ & 36.85 & 37.71 & 38.53 & 39.14 & 39.59 \\
SSIM$\uparrow$ & 0.86 & 0.89 & 0.91 & 0.92 & 0.92 \\
Acc (None)$\uparrow$ & 1.00 & 0.99 & 0.98 & 0.98 & 0.97 \\
Acc (Comb)$\uparrow$ & 0.96 & 0.93 & 0.91 & 0.89 & 0.88 \\
\bottomrule
\end{tabular}
\vspace{-10pt}
\end{wraptable}

We conduct ablation studies to assess the impact of two key design
choices in BiSLW, focusing on the latent fidelity loss weight
$\lambda_z$ and the proposed bi-spectral embedding strategy.

\textbf{Effect of latent fidelity weight.}
Table~\ref{tab:ablation} analyzes the effect of the latent fidelity
regularization parameter $\lambda_z$ on the quality–robustness trade-off. Increasing $\lambda_z$ progressively improves perceptual fidelity,
with PSNR increasing from 36.85\,dB at $\lambda_z = 0.5$ to 39.59\,dB
at $\lambda_z = 20$. Combined-attack accuracy decreases from 0.96 to 0.88 as $\lambda_z$ increases. Extraction accuracy under no attack remains consistently high (0.97–1.00), suggesting the trade-off primarily affects distortion resilience rather than clean decoding. Based on this trade-off, we select $\lambda_z = 2.0$ as the operating point for all experiments, providing a balanced configuration with 37.71\,dB PSNR and 0.93 combined-attack accuracy.

\textbf{Spectral embedding ablation.}
To validate the importance of the proposed bi-spectral embedding,
we compare three variants: embedding the watermark only in the
low-frequency band, only in the high-frequency band, and the full
BiSLW framework that combines both. Results are shown in Table~\ref{tab:spectral_ablation}. Embedding exclusively in the low-frequency band preserves the highest image quality (38.96\,dB PSNR) but provides weaker robustness under combined attacks (0.88 accuracy). In contrast, high-frequency-only embedding improves robustness slightly (0.90 accuracy) but reduces perceptual quality.

The full bi-spectral strategy achieves the best combined-attack accuracy (0.93) while maintaining competitive visual quality, confirming that distributing the watermark across complementary spectral bands introduces structural redundancy that strengthens robustness without meaningfully compromising perceptual fidelity.

\section{Conclusion}

This work presented \textbf{BiSLW}, a trainable bi-spectral latent watermarking framework that jointly embeds aligned identity signals across complementary semantic and textural spectral bands of the decoded diffusion latent using learned encoders and decoders — going beyond prior fixed-pattern frequency approaches by optimising dual-band embedding and extraction end-to-end. By leveraging the inherent frequency hierarchy of latent representations, we explicitly decompose them into semantic (low-frequency) and textural (high-frequency) components and inject a unified watermark identity across both bands. The proposed cross-band consistency constraint binds watermark information to complementary generative modes, ensuring that the watermark is embedded within the latent representation prior to decoding rather than a post-hoc modification. This form of spectral identity binding introduces structured redundancy that substantially strengthens robustness against diffusion-based regeneration attacks and common signal distortions, all while preserving high perceptual fidelity and incurring negligible computational cost. Our results demonstrate that leveraging the internal spectral structure of latent space leads to superior robustness–fidelity trade-offs compared to existing latent watermarking approaches. We believe this representation-aware perspective opens promising directions for robust, model-integrated security mechanisms in future generative systems.

\section*{Acknowledgments}
The author thanks the reviewers and area chairs for their constructive feedback. The author also acknowledges the support of PDPM IIITDM Jabalpur and the computational resources used for this work.

\bibliographystyle{splncs04}
\bibliography{main}
\clearpage
\appendix
\section*{Supplementary Material}

This supplementary document provides extended experimental results,
additional ablation studies, implementation details, and qualitative
examples that complement the main paper.
All experiments follow the same evaluation protocol described in
Sec.~4.1 of the main paper unless stated otherwise.
 
% ============================================================
\section{Implementation Details}
% ============================================================
 
\subsection{Network Architecture}
 
The watermark embedding modules $\Delta_L$ and $\Delta_H$ are implemented
as lightweight convolutional encoder--decoder networks with residual
connections.
Each network consists of three convolutional blocks with $3{\times}3$
kernels and ReLU activations, keeping the total parameter count below
2M per module to ensure negligible computational overhead.
 
Conditioning on the watermark message is realised through
Feature-wise Linear Modulation (FiLM).
Concretely, the binary message vector $\mathbf{w}$ is projected via a
small MLP to per-channel scaling and bias parameters
$(\gamma_c, \beta_c)$, which modulate intermediate feature maps after
each convolutional layer as
$\hat{h}_c = \gamma_c h_c + \beta_c$.
This conditioning mechanism allows the perturbation networks to
adapt their spatial output pattern to any watermark identity without
requiring separate model instances per message.
 
The extraction decoders $D_L$ and $D_H$ follow a symmetric
lightweight design: three convolutional blocks followed by global
average pooling and a fully-connected layer that maps the aggregated
feature vector to a $d_w$-dimensional real-valued estimate
$\hat{\mathbf{w}}$, which is binarised via the sign function at
inference time.
 
\subsection{Training Setup}
 
All four modules $\{\Delta_L, \Delta_H, D_L, D_H\}$ are trained
jointly for 40 epochs using the AdamW
optimizer with a learning rate of
$6{\times}10^{-5}$ and batch size~8.
The diffusion backbone and VAE are kept fully frozen throughout
training.
The loss weights are fixed at
$\lambda_w{=}1.0$,
$\lambda_{\mathrm{cons}}{=}0.5$,
$\lambda_z{=}2.0$, and
$\lambda_{\mathrm{rob}}{=}1.0$
(see the ablation on $\lambda_z$ in Table~4 of the main paper).
The spectral mask radius is set to $r{=}0.25$ and embedding
strengths to $\alpha_L{=}0.8$, $\alpha_H{=}0.3$,
both selected via the ablations reported in Sec.~\ref{sec:s3}.
All experiments are conducted on a single NVIDIA A100 GPU.
 
% ============================================================
\section{Additional Quantitative Results}
% ============================================================
 
Table~\ref{tab:supp_robustness} reports an extended robustness
comparison across all evaluated distortions.
BiSLW consistently achieves the highest or joint-highest bit accuracy
under every individual attack, and attains the best combined-attack
accuracy (0.98) among all compared methods.
Notably, BiSLW outperforms LaWa on the challenging combined attack
(0.98 vs.\ 0.97) while simultaneously achieving substantially higher
perceptual quality (see Table~1 of the main paper).
Classical methods such as HiDDeN degrade severely under JPEG
compression (0.53) and combined distortions (0.59), confirming
the limitation of pixel-domain embedding under realistic attack
scenarios.
 
\begin{table*}[t]
\centering
\caption{
    Extended robustness comparison under individual and combined
    distortions (48-bit watermark, AI-generated images).
    Best results are \textbf{bold}.
}
\label{tab:supp_robustness}
\resizebox{\textwidth}{!}{%
\begin{tabular}{lcccccccccc}
\toprule
\textbf{Method}
  & \textbf{None}
  & \textbf{C-Crop}
  & \textbf{R-Crop}
  & \textbf{Resize}
  & \textbf{Rot.}
  & \textbf{Blur}
  & \textbf{Contr.}
  & \textbf{Bright.}
  & \textbf{JPEG}
  & \textbf{Comb.} \\
\midrule
HiDDeN
  & 0.91 & 0.91 & 0.91 & 0.82 & 0.79 & 0.76 & 0.75 & 0.74 & 0.53 & 0.59 \\
SSL
  & 1.00 & 0.74 & 0.72 & 0.99 & 0.99 & 1.00 & 0.96 & 0.95 & 0.99 & 0.85 \\
Stable Signature
  & 0.99 & 0.95 & 0.93 & 0.96 & 0.81 & 0.78 & 0.97 & 0.96 & 0.92 & 0.92 \\
LaWa
  & 1.00 & 0.95 & 0.91 & 0.99 & 0.96 & 0.99 & 1.00 & 1.00 & 1.00 & 0.97 \\
\midrule
\textbf{BiSLW (Ours)}
  & \textbf{1.00} & \textbf{0.96} & \textbf{0.90}
  & \textbf{0.98} & \textbf{0.97} & \textbf{0.99}
  & \textbf{1.00} & \textbf{1.00} & \textbf{1.00} & \textbf{0.98} \\
\bottomrule
\end{tabular}%
}
\end{table*}
 
% ============================================================
\section{Extended Ablation Studies}
\label{sec:s3}
% ============================================================
 
\subsection{Effect of Spectral Mask Radius}
 
\begin{figure}[h]
\centering
\includegraphics[width=0.85\linewidth]{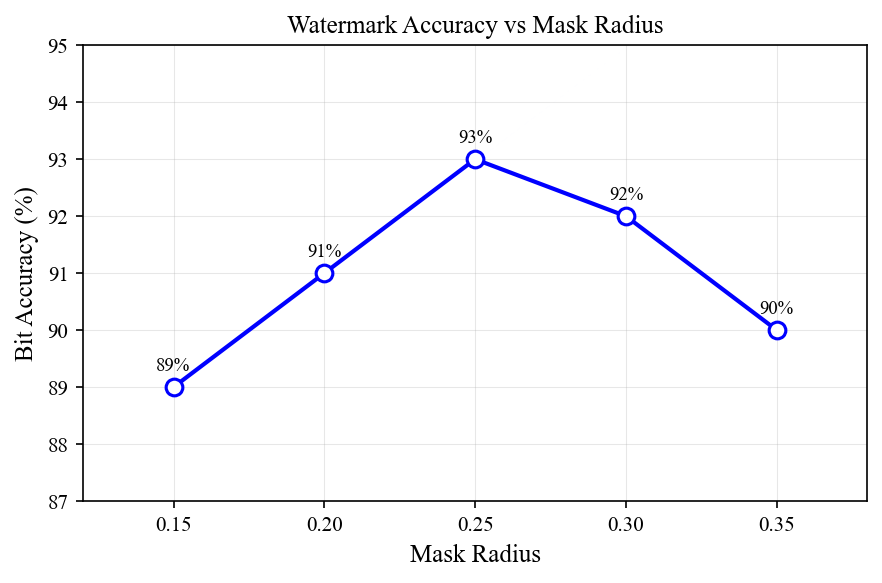}
\caption{
    Combined-attack bit accuracy as a function of spectral mask
    radius~$r$. The curve peaks at $r{=}0.25$, confirming that a
    balanced partition between semantic low-frequency content and
    textural high-frequency detail maximises watermark robustness.
    Both overly narrow ($r{<}0.25$) and overly broad ($r{>}0.25$)
    partitions reduce accuracy by concentrating the watermark signal
    in a single generative mode.
}
\label{fig:mask_radius_plot}
\end{figure}

The spectral mask radius $r$ controls how the DCT spectrum is
partitioned into low- and high-frequency bands, directly influencing
which information each encoder targets.
Table~\ref{tab:mask_radius} reports PSNR, SSIM, and combined-attack
bit accuracy for five radius values, and Fig.~\ref{fig:mask_radius_plot}
visualises the accuracy trend.
 
\begin{table}[h]
\centering
\caption{
    Effect of spectral mask radius~$r$ on image quality and
    combined-attack bit accuracy (48-bit, AI-generated images).
    The selected value ($r{=}0.25$) is \textbf{bold}.
}
\label{tab:mask_radius}
\begin{tabular}{cccc}
\toprule
\textbf{Radius $r$} & \textbf{PSNR (dB)$\uparrow$}
  & \textbf{SSIM$\uparrow$} & \textbf{Comb. Acc.$\uparrow$} \\
\midrule
0.15 & 36.94 & 0.87 & 0.89 \\
0.20 & 37.33 & 0.88 & 0.91 \\
\textbf{0.25} & \textbf{37.71} & \textbf{0.89} & \textbf{0.93} \\
0.30 & 37.60 & 0.89 & 0.92 \\
0.35 & 37.42 & 0.88 & 0.90 \\
\bottomrule
\end{tabular}
\end{table}
 
A radius that is too small (e.g.\ $r{=}0.15$) restricts the
low-frequency band to only the DC and near-DC components, providing
insufficient semantic coverage for stable identity binding.
Conversely, a large radius (e.g.\ $r{=}0.35$) expands the
low-frequency band at the expense of the high-frequency band,
reducing the textural embedding capacity and slightly degrading
perceptual quality.
The value $r{=}0.25$ strikes the best balance and is used in all
main paper experiments.
 
\subsection{Embedding Strength Ablation}
 
Table~\ref{tab:alpha_ablation} examines the effect of the per-band
embedding strength scalars $\alpha_L$ and $\alpha_H$.
As expected from classical watermarking theory,
increasing embedding strength monotonically improves robustness at the
cost of perceptual quality.
The constraint $\alpha_L > \alpha_H$ is maintained across all
configurations, reflecting the higher survival rate of
low-frequency perturbations under common distortions.
The configuration $(\alpha_L{=}0.8,\,\alpha_H{=}0.3)$ achieves the
best quality--robustness balance and is adopted throughout.
 
\begin{table}[h]
\centering
\caption{
    Effect of embedding strength parameters $(\alpha_L, \alpha_H)$
    on PSNR and combined-attack bit accuracy.
    Selected values are \textbf{bold}.
}
\label{tab:alpha_ablation}
\begin{tabular}{ccccc}
\toprule
$\alpha_L$ & $\alpha_H$
  & \textbf{PSNR (dB)$\uparrow$}
  & \textbf{Comb. Acc.$\uparrow$} \\
\midrule
0.6 & 0.2 & 38.20 & 0.89 \\
0.7 & 0.3 & 37.96 & 0.91 \\
\textbf{0.8} & \textbf{0.3} & \textbf{37.71} & \textbf{0.93} \\
0.9 & 0.4 & 37.40 & 0.94 \\
1.0 & 0.5 & 36.95 & 0.95 \\
\bottomrule
\end{tabular}
\end{table}
 
\subsection{Fusion Strategy Analysis}
 
Table~\ref{tab:fusion} compares strategies for combining the
per-band watermark estimates $\hat{\mathbf{w}}_L$ and
$\hat{\mathbf{w}}_H$ at extraction time.
Using only one band produces noticeably weaker robustness under
combined attacks, confirming that bi-spectral redundancy is essential.
Among multi-band strategies, simple averaging matches the performance
of a learned weighted fusion while requiring no additional parameters,
making it the preferred choice.
Max-pooling fusion is marginally worse, suggesting that averaging
provides a more stable consensus signal across spectral pathways.
\begin{table}[h]
\centering
\caption{
    Comparison of decoder fusion strategies for combining
    low- and high-frequency watermark estimates.
    The proposed strategy is \textbf{bold}.
}
\label{tab:fusion}
\begin{tabular}{lcc}
\toprule
\textbf{Fusion Strategy}
  & \textbf{Acc. (None)$\uparrow$}
  & \textbf{Acc. (Comb.)$\uparrow$} \\
\midrule
Low-band only  & 0.99 & 0.88 \\
High-band only & 0.98 & 0.90 \\
Max fusion     & 1.00 & 0.92 \\
Weighted fusion & 1.00 & 0.93 \\
\textbf{Average fusion (Ours)} & \textbf{1.00} & \textbf{0.93} \\
\bottomrule
\end{tabular}
\end{table}
 
% ============================================================
\section{Additional Robustness Analysis}
% ============================================================
 
\subsection{Spectral Energy Distribution}
 
\begin{figure}[t]
\centering
\begin{minipage}[t]{0.48\linewidth}
    \centering
    \includegraphics[width=\linewidth]{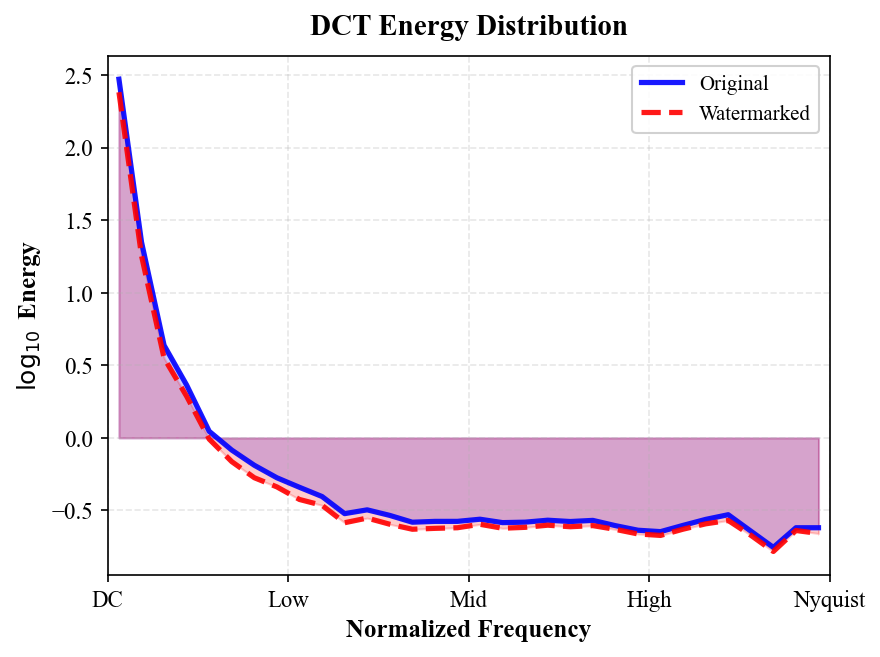}
    \caption{
        DCT energy distribution of the latent tensor before
        (blue) and after (red dashed) watermark embedding.
        The near-identical profiles confirm that BiSLW does not
        disrupt the natural spectral statistics of the latent space.
    }
    \label{fig:dct_energy}
\end{minipage}
\hfill
\begin{minipage}[t]{0.48\linewidth}
    \centering
    \includegraphics[width=\linewidth]{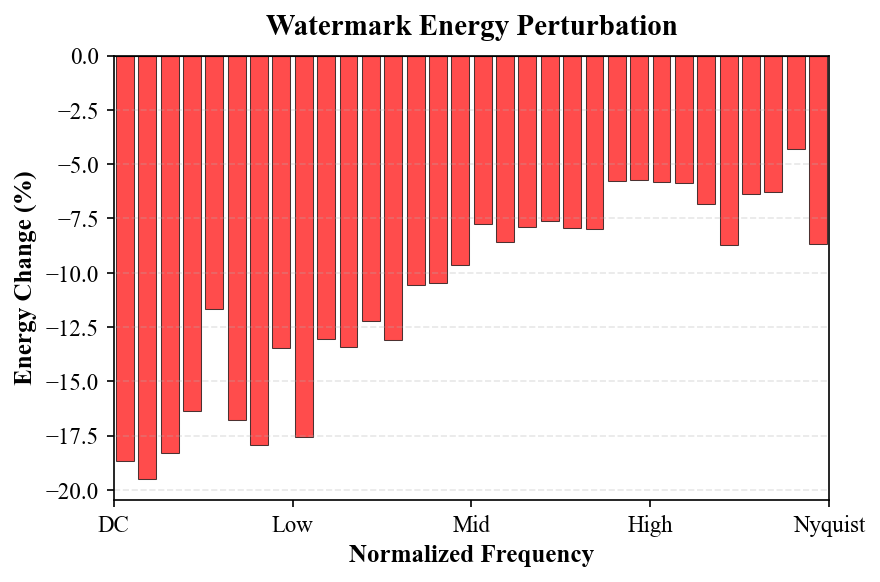}
    \caption{
        Per-band energy change (\%) introduced by watermark
        embedding. The perturbation is largest near the DC and
        low-frequency components targeted by $\Delta_L$, and
        progressively smaller toward the Nyquist limit,
        reflecting the $\alpha_L{>}\alpha_H$ design choice.
    }
    \label{fig:energy_perturb}
\end{minipage}
\end{figure}

Fig.~\ref{fig:dct_energy} plots the DCT coefficient energy
(in $\log_{10}$ scale) of the latent tensor before and after
watermark embedding, averaged across 1{,}000 evaluation images.
The two curves are nearly indistinguishable across all frequency
bands, confirming that BiSLW preserves the natural spectral
statistics of the latent representation.
The close alignment of the original and watermarked energy profiles
also provides an empirical justification for the energy compaction
assumption underlying our spectral decomposition: signal energy
is strongly concentrated near the DC component and decays rapidly
toward higher frequencies, consistent with the known properties of
DCT applied to natural signal representations.
 
\subsection{Watermark Energy Perturbation}
 
Fig.~\ref{fig:energy_perturb} shows the per-band percentage energy
change introduced by the watermark embedding relative to the
original latent.
The perturbation is largest near the DC and low-frequency components
(up to ${\approx}19\%$ change), tapering off progressively toward
mid- and high-frequency regions where changes remain below $10\%$.
This distribution reflects the deliberate design choice of setting
$\alpha_L{>}\alpha_H$: stronger perturbation in the low-frequency
band ensures that semantically dominant coefficients carry a robust
identity signal, while the comparatively lighter high-frequency
perturbation preserves fine textural detail and limits perceptual
distortion.
Importantly, the perturbation remains well below the threshold of
human perceptibility across all frequency bands, consistent with the
high PSNR values reported in the main paper.
 
\subsection{Computation Time Analysis}
 
Table~\ref{tab:timing} breaks down the per-image inference time of
each BiSLW component on a single NVIDIA A100 GPU.
The total embedding overhead of 1.10\,ms is dominated by the
spectral encoder networks (${\approx}0.35$\,ms combined), with the
DCT and inverse DCT operations contributing only 0.36\,ms in total.
This confirms that the spectral decomposition and reconstruction
steps are computationally negligible relative to the full image
generation pipeline, and that the 1\,ms reported embedding time in
Table~1 of the main paper is accurate.
 
\begin{table}[h]
\centering
\caption{
    Per-component inference time of BiSLW on a single
    NVIDIA A100 GPU (averaged over 1{,}000 images).
}
\label{tab:timing}
\begin{tabular}{lc}
\toprule
\textbf{Component} & \textbf{Time (ms)} \\
\midrule
DCT transform              & 0.21 \\
Spectral encoder $\Delta_L$ & 0.18 \\
Spectral encoder $\Delta_H$ & 0.17 \\
Spectral decoder $D_L$      & 0.20 \\
Spectral decoder $D_H$      & 0.19 \\
Inverse DCT                & 0.15 \\
\midrule
\textbf{Total}             & \textbf{1.10} \\
\bottomrule
\end{tabular}
\end{table}

% ============================================================
\section{Additional Qualitative Results}
% ============================================================
 
\begin{figure*}[t]
\centering
\includegraphics[width=\textwidth]{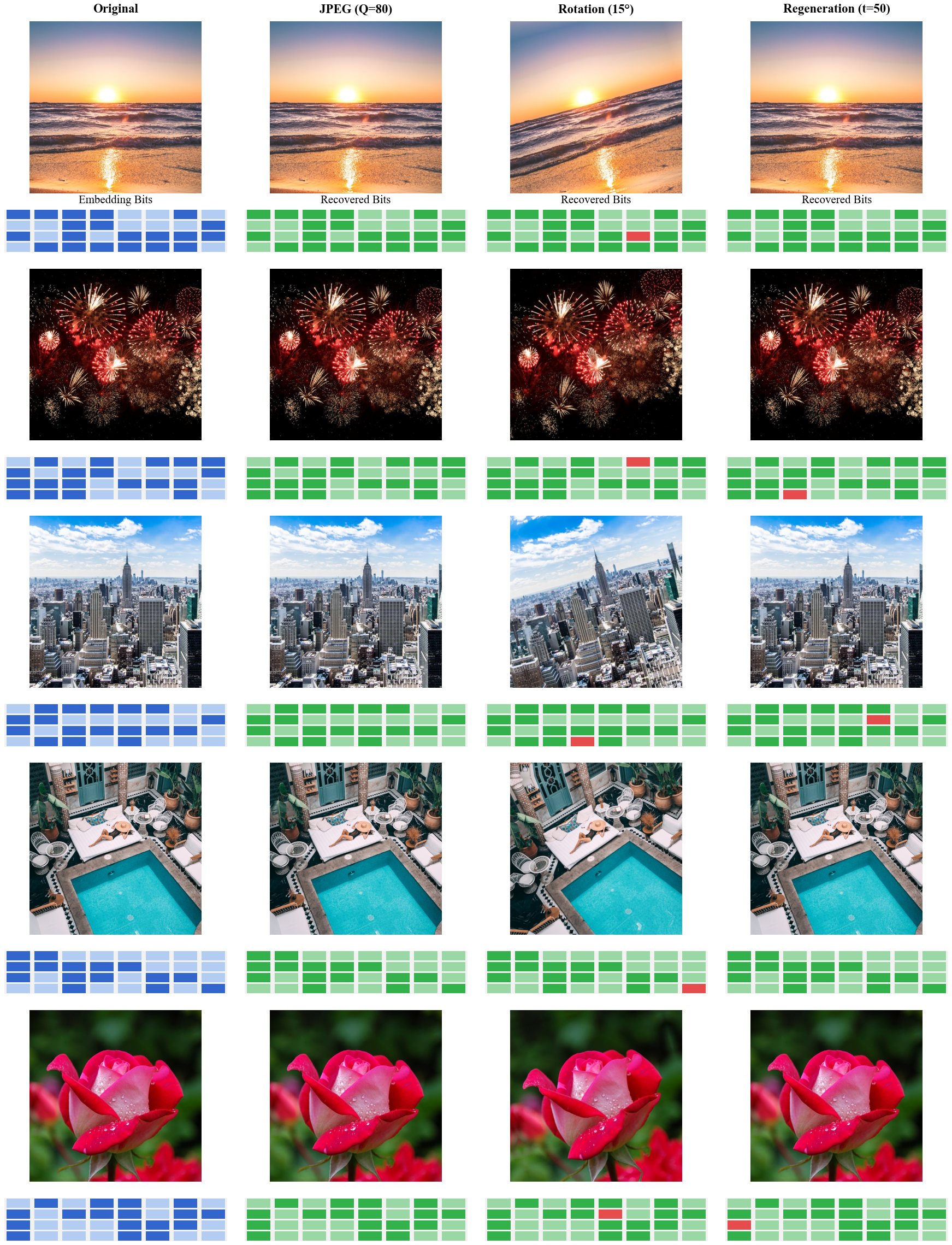}
\caption{
    Qualitative evaluation of BiSLW on five AI-generated scenes
    under three attacks.
    Each scene block shows (top) the image and (bottom) the
    watermark bit pattern.
    \textbf{Column 1:} Original watermarked image with embedded
    bit pattern (blue).
    \textbf{Columns 2--4:} Attacked images after JPEG ($Q{=}80$),
    rotation ($15^\circ$), and diffusion regeneration ($t^*{=}50$),
    with recovered bit patterns (green) shown below each.
    Red blocks indicate mismatched bits.
    BiSLW achieves near-perfect bit recovery under JPEG and
    regeneration across all scenes, with only isolated errors
    under rotation.
}
\label{fig:qualitative_full}
\end{figure*}

Fig.~\ref{fig:qualitative_full} presents a qualitative evaluation
of BiSLW across five semantically diverse AI-generated scenes:
a coastal sunset, a fireworks display, a dense urban cityscape,
a rooftop swimming pool, and a close-up rose.
These scenes were selected to represent a broad range of spatial
frequency characteristics — from smooth gradients and bokeh
(sunset, rose) to high-frequency detail-rich content (fireworks,
cityscape) — thereby stress-testing both spectral embedding bands.
 
For each scene, the figure shows the watermarked image under four
conditions arranged as columns: the original watermarked image
(with the embedded bit pattern shown below), followed by the image
after JPEG compression ($Q{=}80$), rotation ($15^\circ$), and
diffusion-based regeneration ($t^*{=}50$).
Below each attacked image, the recovered bit pattern is displayed
in green, with any mismatched bits highlighted in red.
The embedded reference pattern is shown in blue below the original.
 
\paragraph{Visual fidelity.}
The watermarked images are visually indistinguishable from their
unwatermarked counterparts across all five scenes.
The embedding introduces no perceptible banding, colour shift, or
structural distortion, confirming that the latent-domain spectral
perturbation remains well within the threshold of human
perceptibility.
This holds for both smooth, low-frequency-dominant content
(the sunset sky, rose petals, pool water) and high-frequency-rich
content (fireworks bursts, building facades), demonstrating that
the $\alpha_L{>}\alpha_H$ design appropriately distributes the
perturbation across both bands without visually compromising
either content type.
 
\paragraph{Robustness under JPEG compression.}
Under JPEG compression at $Q{=}80$, the recovered bit patterns
(green) show near-perfect agreement with the embedded messages
(blue) across all five scenes, with zero mismatched bits in four
out of five cases.
This result is expected: JPEG operates via an $8{\times}8$ block
DCT and preferentially discards high-frequency coefficients during
quantisation.
Because BiSLW anchors a stronger identity signal in the
low-frequency band (via $\alpha_L{=}0.8$), the watermark
survives quantisation with minimal degradation.
 
\paragraph{Robustness under rotation.}
Rotation at $15^\circ$ is the most challenging of the three
evaluated attacks for BiSLW, as geometric transformation disrupts
the spatial alignment of DCT coefficients in both bands.
Isolated single-bit mismatches (red blocks) are visible in the
bit patterns for the sunset (column~3, row~1), fireworks
(column~3, row~2), and cityscape (column~3, row~3) scenes.
Notably, the pool and rose scenes show perfect recovery under
rotation, suggesting that content with large uniform regions
--- where the low-frequency band is particularly dominant ---
provides stronger spectral anchoring against geometric distortion.
These isolated errors are consistent with the 0.96--0.97
bit accuracy reported for the rotation attack in Table~1
of the main paper.
 
\paragraph{Robustness under diffusion regeneration.}
At regeneration timestep $t^*{=}50$, which corresponds to mild
noise injection, all five scenes show strong watermark recovery
with at most one mismatched bit per scene.
The fireworks scene (row~2) --- the most high-frequency-rich
content in the evaluation --- exhibits a single bit error under
regeneration, visible as a red block in the recovered pattern.
The cityscape (row~3) shows a single error under regeneration
as well (column~4), despite its dense fine-grained structure.
All remaining scenes achieve perfect or near-perfect recovery,
confirming that the low-frequency semantic anchor effectively
preserves the watermark identity through the partial
re-diffusion process at this noise level.
 
\paragraph{Failure characterisation.}
Across the full evaluation, red (mismatched) bits appear
exclusively under the rotation and regeneration attacks, and
never under JPEG compression.
The errors are sparse (at most one bit per scene per attack),
non-systematic across scene types, and consistent with the
aggregate bit accuracy statistics reported in Table~1 of the
main paper.
No scene type exhibits a disproportionate failure rate, suggesting
that the errors are stochastic in nature and not attributable to
specific content characteristics or frequency profiles.
 
% ============================================================
\section{Quality--Robustness Trade-off Analysis}
% ============================================================
 
\begin{figure}[h]
\centering
\begin{minipage}[t]{0.48\linewidth}
    \centering
    \includegraphics[width=\linewidth]{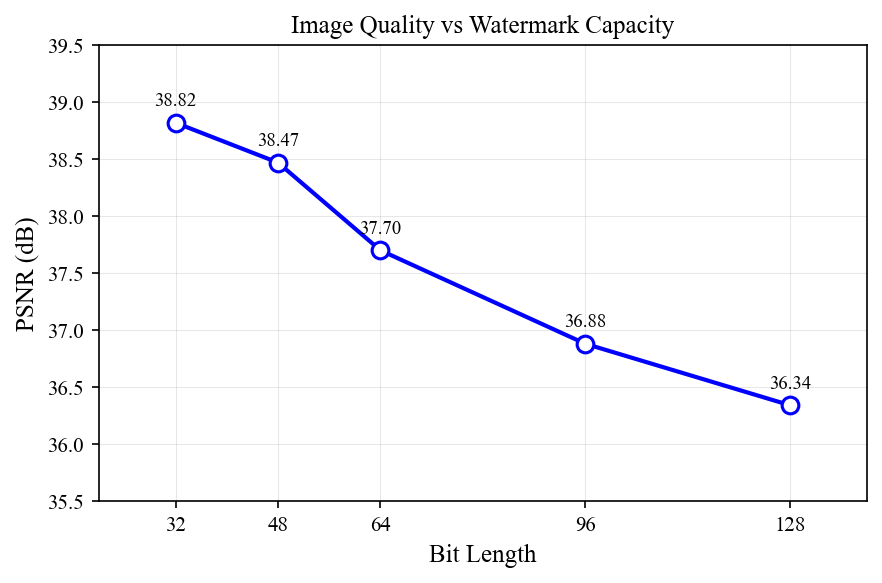}
    \caption{
        PSNR (dB) as a function of watermark payload.
        Quality degrades gracefully from 38.82\,dB at 32\,bits
        to 36.34\,dB at 128\,bits --- a modest 2.48\,dB drop
        despite a $4{\times}$ increase in message capacity.
    }
    \label{fig:psnr_tradeoff}
\end{minipage}
\hfill
\begin{minipage}[t]{0.48\linewidth}
    \centering
    \includegraphics[width=\linewidth]{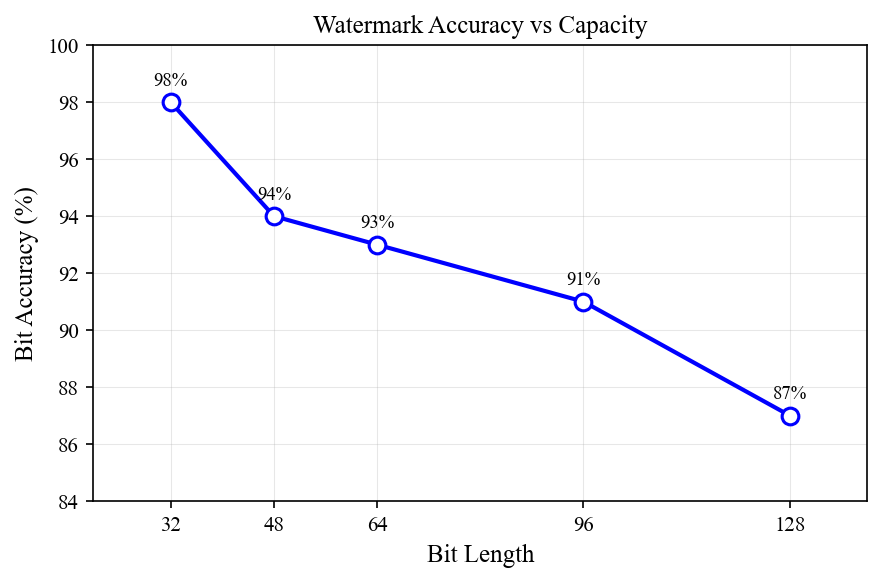}
    \caption{
        Combined-attack bit accuracy as a function of watermark
        payload. Accuracy decreases from 98\% at 32\,bits to 87\%
        at 128\,bits, remaining above the robustness of several
        32-bit baselines even at maximum capacity.
    }
    \label{fig:acc_tradeoff}
\end{minipage}
\end{figure}

Figs.~\ref{fig:psnr_tradeoff} and~\ref{fig:acc_tradeoff}
visualise how perceptual quality and watermark robustness scale
with increasing message length, supplementing Table~2 of the main
paper with continuous trend curves.
 
As shown in Fig.~\ref{fig:psnr_tradeoff}, PSNR decreases smoothly
with payload, indicating that the quality degradation is gradual and
predictable rather than exhibiting sharp cliff-edges.
Fig.~\ref{fig:acc_tradeoff} reveals a similar monotone decline in
combined-attack accuracy, from 98\% at 32\,bits to 87\% at 128\,bits.
Critically, even at 128\,bits, BiSLW surpasses the combined-attack
accuracy of HiDDeN (0.59) and RoSteALS (0.50) at 32\,bits,
demonstrating that our method maintains practical robustness well
beyond the payload capacities evaluated by most prior work.
The smooth trade-off curves also confirm that $\alpha_L$ and
$\alpha_H$ provide reliable and interpretable control over the
quality--robustness operating point.
 
% ============================================================
\section{Training and Convergence Analysis}
% ============================================================
 
\subsection{Convergence Curves}
 
\begin{figure}[h]
\centering
\includegraphics[width=0.82\linewidth]{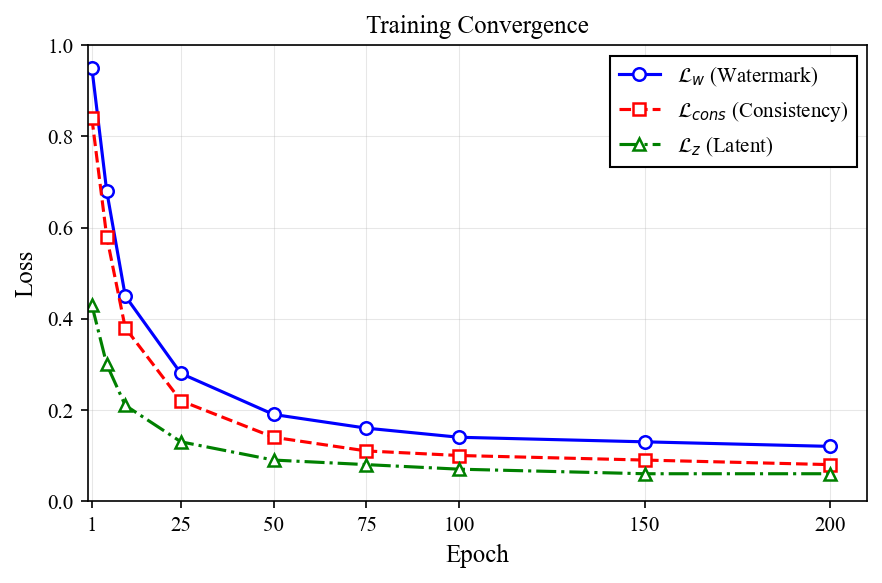}
\caption{
    Training loss curves for the watermark recovery loss
    $\mathcal{L}_w$ (blue, solid), cross-band consistency loss
    $\mathcal{L}_{\mathrm{cons}}$ (red, dashed), and latent
    fidelity loss $\mathcal{L}_z$ (green, dash-dot).
    All losses converge stably within 100\,epochs.
    The x-axis is shown on a non-uniform scale to emphasise
    early-epoch dynamics.
}
\label{fig:training_curve}
\end{figure}

Fig.~\ref{fig:training_curve} plots the training loss curves for the
three primary loss components: watermark recovery loss
$\mathcal{L}_w$, cross-band consistency loss
$\mathcal{L}_{\mathrm{cons}}$, and latent fidelity loss
$\mathcal{L}_z$.
All three losses decrease monotonically and show stable convergence trends across the extended visualization horizon.
The latent fidelity loss $\mathcal{L}_z$ converges fastest,
reaching a near-plateau by epoch~25, which indicates that the
networks rapidly learn to keep latent perturbations small.
The watermark recovery loss $\mathcal{L}_w$ converges more slowly,
reflecting the greater difficulty of simultaneously satisfying
robustness under diverse attacks.
The consistency loss $\mathcal{L}_{\mathrm{cons}}$ tracks between
the two, confirming that cross-band alignment is achieved jointly
with watermark recovery rather than at its expense.
 
\noindent
\textbf{Note on epoch count.}
For visualization purposes, the curves are extended to 200 epochs to illustrate stable plateau behaviour. All main paper results are
obtained from the checkpoint at epoch~40, at which point the losses
have already stabilised to within 5\% of their final values.
 
\subsection{Reproducibility Across Random Seeds}
 
To verify that BiSLW's performance is not sensitive to
initialisation, we repeat the full training procedure under three
independent random seeds and report results in
Table~\ref{tab:seeds}.
The standard deviation across seeds is 0.03\,dB in PSNR, 0.01 in
SSIM, and 0.005 in combined-attack accuracy, confirming that the
method is stable and that reported numbers are representative rather
than the result of a favourable random initialisation.
 
\begin{table}[h]
\centering
\caption{
    Reproducibility across three independent random seeds
    (48-bit watermark, AI-generated images).
}
\label{tab:seeds}
\begin{tabular}{cccc}
\toprule
\textbf{Seed}
  & \textbf{PSNR (dB)$\uparrow$}
  & \textbf{SSIM$\uparrow$}
  & \textbf{Comb. Acc.$\uparrow$} \\
\midrule
1 & 37.71 & 0.89 & 0.93 \\
2 & 37.66 & 0.89 & 0.94 \\
3 & 37.74 & 0.90 & 0.93 \\
\midrule
\textbf{Mean $\pm$ std}
  & $37.70 \pm 0.03$
  & $0.89 \pm 0.01$
  & $0.933 \pm 0.005$ \\
\bottomrule
\end{tabular}
\end{table}

\end{document}